\RequirePackage{fix-cm}
\documentclass[twocolumn]{svjour3}          
\smartqed  
\usepackage{graphicx}
\usepackage{hyperref}       
\usepackage{url}            
\usepackage{booktabs}       
\usepackage{amsfonts}       
\usepackage{nicefrac}       
\usepackage{microtype}      
\usepackage{subfig}
\usepackage{multirow}
\usepackage{multicol}
\usepackage{amssymb}
\usepackage{amsmath}

\usepackage[dvipsnames]{xcolor}

\newcommand{\red}[1]{\textcolor{black}{#1}}
\newcommand{\blue}[1]{\textcolor{black}{#1}}
\begin{document}

\title{Exploring the Semi-supervised Video Object Segmentation Problem from a Cyclic Perspective}


\author{Yuxi Li \and 
        Ning Xu \and 
        Wenjie Yang \and
        John See \and
        Weiyao Lin
}


\institute{
            Yuxi Li \at
              Department of Electronic Engineering, Shanghai Jiao Tong University, Shanghai China \\
              \email{lyxok1@sjtu.edu.cn}
           \and
           Ning Xu \at
              Adobe Research, San Jose, USA \\
              \email{nxu@adobe.com}
            \and
            Wenjie Yang \at
            Department of Computer Science, Shanghai Jiao Tong University, Shanghai China \\
              \email{13633491388@sjtu.edu.cn}
            \and
            John See \at
              Heriot-Watt University, Malaysia \\
              \email{J.See@hw.ac.uk}
           \and
           Weiyao Lin \at
              Department of Electronic Engineering, Shanghai Jiao Tong University, Shanghai China \\
              \email{wylin@sjtu.edu.cn}
}

\date{Received: date / Accepted: date}

\maketitle

\begin{abstract}
  Modern video object segmentation (VOS) algorithms have achieved remarkably high performance in a sequential processing order, while most of currently prevailing pipelines still show some obvious inadequacy like accumulative error, unknown robustness or lack of proper interpretation tools. In this paper, we place the semi-supervised video object segmentation problem into a cyclic workflow and find the defects above can be collectively addressed via the inherent cyclic property of semi-supervised VOS systems. Firstly, a cyclic mechanism incorporated to the standard sequential flow can produce more consistent representations for pixel-wise correspondance. Relying on the accurate reference mask in the starting frame, we show that the error propagation problem can be mitigated. Next, a simple gradient correction module, which naturally extends the offline cyclic pipeline to an online manner, can highlight the high-frequent and detailed part of results to further improve the segmentation quality while keeping feasible computation cost. Meanwhile such correction can protect the network from severe performance degration resulted from interference signals. Finally we develop cycle effective receptive field (cycle-ERF) based on gradient correction process to provide a new perspective into analyzing object-specific regions of interests. We conduct comprehensive comparison and detailed analysis on challenging benchmarks of DAVIS16, DAVIS17 and Youtube-VOS, demonstrating that the cyclic mechanism is helpful to enhance segmentation quality, improve the robustness of VOS systems, and further provide qualitative comparison and interpretation on how different VOS algorithms work. The code of this project can be found at \url{https://github.com/lyxok1/STM-Training}. \footnote{
            This manuscript is an extended version of our conference paper to be published at the Thirty-fifth Conference on Neural Information Processing Systems (NeurIPS) 2020. \emph{delving into the cyclic mechanism of semi-supervised video object segmentation}~\cite{Li_2020_NeurIPS}. We have cited this paper in the manuscript and extended the paper substantially but not limited in following aspects: (1). A smooth regularized term and insight from frequency domain is appended in the method part. (2). In depth analysis on the robust of VOS model and the effect from our correction methods is included in the methods part. (3) More comprehensive experiments are appended to demonstrate the generality of our studies, including comparison under different baseline models and backbones, results with COCO pretraining and more qualitative results of effect from core components.
}
\end{abstract}

\section{Introduction}
Video object segmentation (VOS) is garnering more attention in recent years due to its widespread application in the area of video editing and analysis. Among all the VOS scenarios, semi-supervised video object segmentation is the most practical and widely researched. Specifically, a mask is provided in the first frame indicating the location and boundary of the objects, and the algorithm should accurately segment the same objects from the background in subsequent frames. A natural solution toward this problem is to process videos in a sequential order, exploiting the information from previous frames and guides the segmentation process in the current frame, since in most practical scenarios, the video is obtained in an online manner where only previous knowledge is available. Following this manner, current state-of-the-ar pipelines~\cite{Oh_2018_CVPR,Lin_2019_ICCV,luiten2018premvos,Oh_2019_ICCV,Voigtlaender_2019_CVPR,Zeng_2019_ICCV,KMN,GCVOS2020} achieve high segmentation quality by delving into the information reuse from previous frames. 

\begin{figure*}
    \centering
    \includegraphics[width=\textwidth]{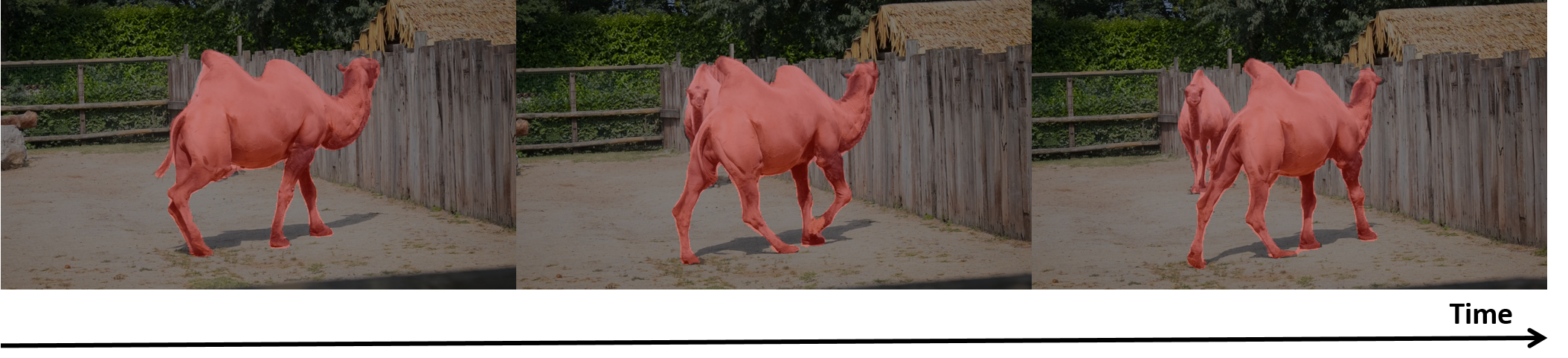}
    \caption{An example of error propagation risk during the inference time, while the reference object is the camel in foreground, the distracting camel from background is incorrectly segmented at the same time.}
    \label{fig:error}
\end{figure*}

However, few research consider the flaws exposed by such sequential processing paradigm, where currently prevailing VOS pipelines still exhibit the following problems: (1) \textbf{Prone to accumulative error}. Ideally, if the masks predicted for intermediate frames are sufficiently accurate, they can provide more helpful object-specific information. Nevertheless, erroneous intermediate masks can mislead the segmentation procedure in future frames (as exemplified in Figure~\ref{fig:error}), and this error is further enlarged when low-quality segmentation dominate the reference templates. 
(2) \textbf{Robustness}. Although there is no research explicitly analyzing the robustness of VOS systems, but intuitively, it can be easily affected by noise manually appended into the previous knowledge, in cases such as adversarial attacks~\cite{goodfellow2014explaining} particularly aimed at VOS algorithms. (3) \textbf{Unavailable tools for interpretation.} There is no unified tool to generally show how a VOS network is working given input frame and reference knowledge.

 In this paper, we consider the problems above in a cyclical context and find the cyclic mechanism can be a potentially unified solution to collectively address these issues. Semi-supervised VOS is inherently suitable to be combined with a cyclic manner. Different from the predicted reference masks, the initial reference mask provided in the starting frame is always perfectly accurate and reliable. This inspires us to explicitly bridge the relationship between the initial reference mask and objective frame by taking the first reference mask as a measurement of prediction.
 
Specifically, when applying a generalized forward-backward data flow to form a cyclical structure and training our segmentation network at both the objective frame and starting frame, our model can learn more consistent correspondence relationship between predictions and the initial template mask. Further, at the inference stage, such cyclic structure can be naturally extended to an online version via a gradient correction module, which selectively refines the detailed and high-frequent part of predicted mask based on the gradient backward from the starting frame at a marginal time cost. As a results, the cyclic workflow can effectively suppress the accumulative error with the starting frame as correct measurement. Meanwhile the online correction can also prevent the network from interference of noisy reference, boosting the robustness of our pipeline. Additionally, inspired by the process of gradient correction, we develop a new interpretation tool called cycle effective receptive field (cycle-ERF), which gradually updates an empty objective mask to show the strong response area w.r.t. the reference mask. In our experiments, we utilize the cycle-ERF to analyze how the cyclic training scheme affects the support regions of objects and highlight the difference in focus-area among distinctive baseline methods. This visualization method provides a fresh perspective for analyzing how the segmentation network extracts regions of interests from guidance masks.
 
The trained models are evaluated in both online and offline schemes on common object segmentation benchmarks: DAVIS16~\cite{Perazzi2016}, DAVIS17~\cite{Pont-Tuset_arXiv_2017} and Youtube-VOS~\cite{Xu_2018_S2S_ECCV}, where we combine our cyclic methods to other baseline models and achieve results that are competitive to other state-of-the-art methods under a fair comparison setting. Besides, we also make detailed and comprehensive analysis to show how each part of the cyclic mechanism works under our design.
 
 In a nutshell, the contributions of this paper can be summarized as follows:
 
 \begin{itemize}
     \item We incorporate cycle consistency into the training process of a semi-supervised video object segmentation network to mitigate the error propagation problem and further improve the segmentation quality. We achieved competitive results without data pretraining on mainstream benchmarks and can further be improved with more synthetic data.
     \item We design a gradient correction module to extend the offline segmentation network to an online approach, which boosts the model performance with marginal increase in computation cost, while keeping the model robust to disturbing noise.
     \item We develop cycle-ERF, a new visualization method to analyze the important regions on different segmentation models, which offers interpretability on the impact of cyclic training.
 \end{itemize}

\section{Related works}
\subsection{Semi-supervised video object segmentation}
 Semi-supervised video object segmentation has been widely researched in recent years with the rapid development of deep learning techniques. Depending on the presence of a learning process during inference stage, the segmentation algorithms can be generally divided into \emph{online} methods and \emph{offline} methods. OVOS~\cite{Cae_OVOS_17} is the first online approach to exploit deep learning for the VOS problem, where a multi-stage training strategy is design to gradually shrink the focus of network from general objects to the one in reference masks. Subsequently, OnAVOS~\cite{voigtlaender17BMVC} improved the online learning process with an adaptive mechanism. MaskTrack ~\cite{Perazzi_2017_CVPR} introduced extra static image data with mask annotation and employed data synthesized through affine transformation, to fine-tune the network before inference. All of these online methods require explicit parameter updating during inference. Although high performance can be achieved, these methods are usually time-consuming with a real-time FPS of less than 1, rendering them unfeasible for practical deployment.

On the other hand, there are a number of offline methods that are deliberately designed to learn generalized correspondence feature and they do not require necessary online learning process during inference time. RGMP~\cite{Oh_2018_CVPR} designed an hourglass structure with skip connections to predict the objective mask based on the current frame and previous information. S2S~\cite{Xu_2018_S2S_ECCV} proposed to model video object segmentation as a sequence-to-sequence problem and proceeds to exploit a temporal modeling module to enhance the temporal coherency of mask propagation. Other works like~\cite{Zeng_2019_ICCV,luiten2018premvos} resorted to using state-of-the-art instance segmentation or tracking pipeline~\cite{He2018Mask,peng2020ctracker} while attempting to design matching strategies to associate the mask over time. A few recent methods FEELVOS~\cite{Voigtlaender_2019_CVPR} and AGSS-VOS~\cite{Lin_2019_ICCV} mainly exploited the guidance from the initial reference and the last previous frame to enhance the segmentation accuracy with deliberately designed feature matching scheme or attention mechanism. STM~\cite{Oh_2019_ICCV} further optimized the feature matching process with external feature memory and an attention-based matching strategy, such memorial structure is further optimized by involving global context~\cite{GCVOS2020}, adaptive gaussian kernel~\cite{KMN} and \red{structure of vision transformer~\cite{AOT2021}}. Compared with online methods, these offline approaches are more efficient. However, to learn more general and robust pixel-wise feature correspondence, these data-hungry methods may require backbones pretrained on large amounts of extra data with mask annotations from other tasks such as instance segmentation~\cite{COCO,Everingham15} or saliency detection~\cite{CSSD}. Without these auxiliary help, the methods might well be disrupted by distractions from similar objects in the video, which then propagates erroneous mask information to future frames.

All the approaches mentioned above follow a standard sequential processing order from start to end of the video and can not ensure the predicted mask is closely related to the initial reference guidance at first frame. In contrast, by embedding the cyclic mechanism into training stage, our method explicitly impose the constraint of reference mask in learning process. Besides, the online extension of gradient correction does not update the pretrained model parameters as other online learning methods, but dynamically refine the output according to the modification information from the reliable initial reference mask.

\subsection{Cycle consistency}
Cycle consistency is widely researched in unsupervised and semi-supervised representation learning, where a transformation and its inverse operation are applied sequentially on input data, the consistency requires that the output representation should be close to the original input data in feature space. With this property, cycle consistency can be applied to different types of correspondence-related tasks.\red{~\cite{trackcolor} is a classical technique for correspondence learning, which treat the learning problem as video colorization and impose the consistency on natural color space to force embeddings of the same semantics to be closer in feature space. In the work of~\cite{cycle3d}, a multiple step cycle-loop is build to connect correspondence relationship between real images and rendered shots from 3D models.} ~\cite{CVPR2019_CycleTime} combined patch-wise consistency with a weak tracker to construct a forward-backward data loop and this guides the network to learn representative feature across different time spans,~\cite{jabri2020walk} further extend such cyclic data loop to a random walk process.~\cite{Meister:2018:UUL} exploited the cycle consistency in unsupervised optical flow estimation by designing a bidirectional consensus loss during training. On the other hand Cycle-GAN~\cite{CycleGAN2017} and Recycle-GAN~\cite{Bansal2018Recycle} and other popular examples of how cyclic training can be utilized to learn non-trivial cross-domain mapping, yielding reliable image-to-image transformation across different domains.

Our method with cyclic mechanism is different from the works mentioned above in following aspects. \red{First, the motivation to exploit cycle consistency in our work is to explicitly regularize the predicted mask to be accurate for backward reference in cyclic loop, while in unsupervised methods like~\cite{CVPR2019_CycleTime}, the cycle consistency is an approach to obtain correspondence ground-truth.} \red{Further, the learning objective of our work is still a fully supervised segmentation task, with high-level and clear semantic information (the object is taken as foreground while the others are background). In contrast, methods with unsupervised cycle consistency for correspondence learning construct their objective by self-mimic in low-level semantics (e.g. the color space~\cite{trackcolor}, spatial position~\cite{CVPR2019_CycleTime} or transformation flow~\cite{cycle3d}), thus the learned embeddings are easy to correspond to distractors with similar low-level expression if there is not sufficient context provided. Consequently large amount of data are required to train these unsupervised methods to learn to catch the context relationship.} Finally, our cyclic structure is not only applicable during training, but also useful in the inference stage. By measuring the consistency between initial reference mask and predicted results, we can refine the output on current frame to obtain more accurate guidance for future prediction.

\section{Methods}

\begin{figure*}
    \centering
    \includegraphics[width=\textwidth]{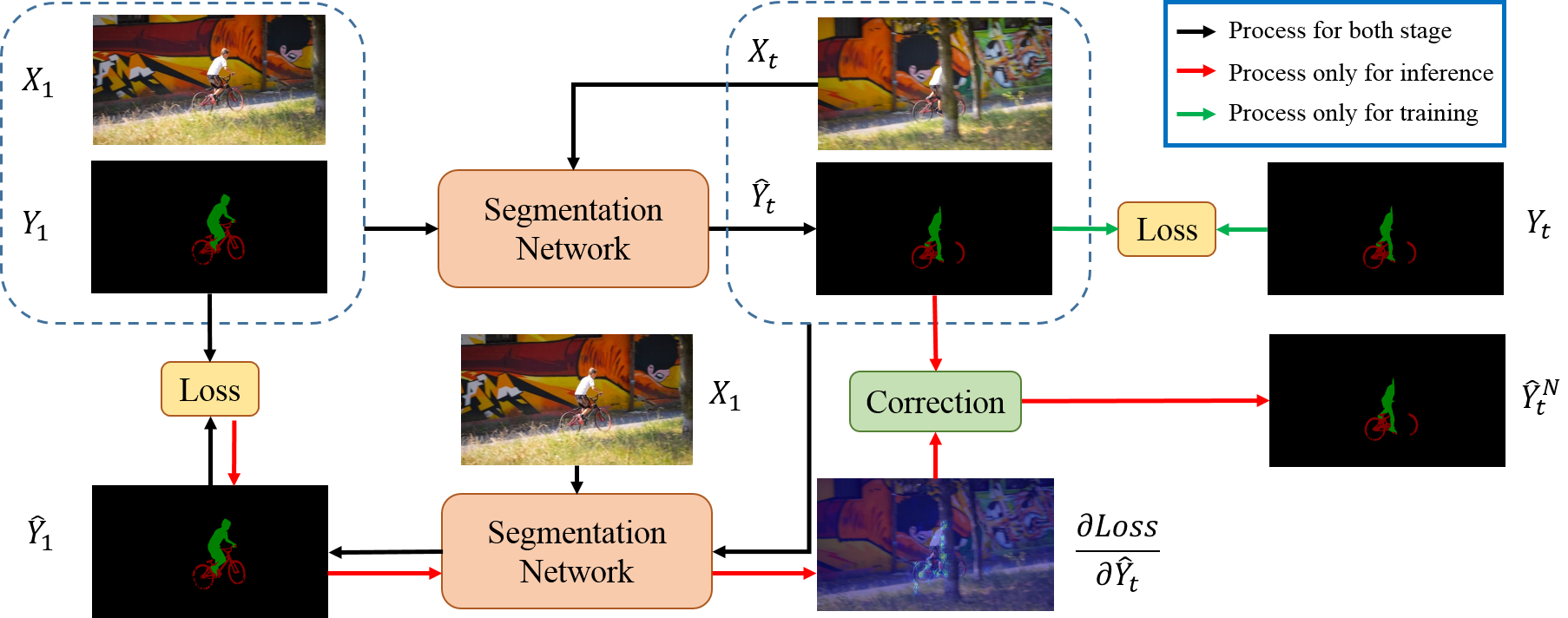}
    \caption{Overview of the proposed cyclic mechanism in both training and inference stages of the segmentation network. For simplicity, we take the situation where $\mathcal{X}_{t-1}=\{X_1\}$, $\mathcal{Y}_{t-1}=\{Y_1\}$, $\widehat{\mathcal{X}}_t=\{X_t\}$ and $\widehat{\mathcal{Y}}_t=\{\widehat{Y}_t\}$ as an example.}
    \label{fig:framework}
\end{figure*}

\subsection{Problem formulation}
Given a video of length $T$, $X_t$ is the $t$-th frame ($t \in [1, T]$) in temporal sequential order, and $Y_t$ is its corresponding annotation mask. $\mathcal{S}_{\theta}$ is an object segmentation network parameterized by learnable weights $\theta$. In terms of the sequential processing order of the video, the segmentation network should achieve the function as in Equation~(\ref{eq:problem}) below:
\begin{equation}\label{eq:problem}
    \widehat{Y}_t = \mathcal{S}_{\theta}\left(\mathcal{X}_{t-1}, \mathcal{Y}_{t-1}, X_t \right) \quad t \in [2, T]
\end{equation}
where $\widehat{Y}_t$ denotes the predicted object mask at $t$-th frame. $\mathcal{X}_{t-1} \subset \{X_i | i \in [1, t-1]\}$ is the reference frame set, which is a subset of all frames appearing before objective frame $X_t$. Similarly, $\mathcal{Y}_{t-1}$ is a set containing reference object masks corresponding to the reference frames in $\mathcal{X}_{t-1}$. However, in the semi-supervised setting, only the initial reference mask at the first frame is available. Therefore, in the inference stage, the corresponding predicted mask $\widehat{Y}_t$ is taken as the approximation of the reference mask. Hence, we have $\mathcal{Y}_{t-1} \subset \{Y_1\}\bigcup\{\widehat{Y}_i | i \in [2, t-1]\}$.

\subsection{Cycle consistency loss}\label{sec:cycle-loss}

For the sake of mitigating error propagation during training, we incorporate the cyclical process into the offline training process to explicitly bridge the relationship between the initial reference and predicted masks. To be specific, as illustrated in Figure~\ref{fig:framework}, after obtaining the predicted output mask $\widehat{Y}_t$ at frame $t$, we construct a \textbf{cyclic reference set} for frames and mask set, respectively denoted as $\widehat{\mathcal{X}}_t \subset \{X_i | i \in [2, t]\}$, $\widehat{\mathcal{Y}}_t \subset \{\widehat{Y}_i | i \in [2, t]\}$.

With the cyclic reference set, we can obtain the prediction for the initial reference mask in the same manner as sequential processing:
\begin{equation}\label{eq:cycle}
    \widehat{Y}_1 = \mathcal{S}_{\theta}\left(\widehat{\mathcal{X}}_{t}, \widehat{\mathcal{Y}}_{t}, X_1 \right)
\end{equation}
Consequently, we apply mask reconstruction loss (in Equation~\ref{eq:cycle-loss}) during supervision, optimizing on both the output mask of the $t$-th frame $\widehat{Y}_t$ and the backward prediction $\widehat{Y}_1$.
\begin{equation}\label{eq:cycle-loss}
    \mathcal{L}_{cycle, t} = \mathcal{L}(\widehat{Y}_t, Y_t) + \mathcal{L}(\widehat{Y}_1, Y_1)
\end{equation}
In implementation, we utilize the combination of cross-entropy loss and mask IOU loss as supervision at both sides of the cyclic loop, which can be formulated as,

\begin{gather}\label{eq:loss}
    \mathcal{L}(\widehat{Y}_t, Y_t) = \mathcal{L}_{IOU} + \gamma\mathcal{L}_{CE} \\
    \mathcal{L}_{IOU} = 1 - \frac{\sum_{u\in \Omega}{\min(\widehat{Y}_{t,u}, Y_{t,u})}}{\sum_{u\in \Omega}{\max(\widehat{Y}_{t,u}, Y_{t,u})}} \\
    \mathcal{L}_{CE} = \sum_{u \in \Omega}{\left(\left(1-Y_{t,u}\right)\log(1-\widehat{Y}_{t,u}) + Y_{t,u}\log(\widehat{Y}_{t,u}) \right)}
\end{gather}
where $\Omega$ denotes the set of all pixel coordinates in the mask while $Y_{t,u}$ and $\widehat{Y}_{t,u}$ are the normalized pixel values at coordinate $u$ of the masks, $\gamma$ is a hyperparameter to balance between the two loss terms. It should also be noted that the cyclic mechanism in Figure~\ref{fig:framework} indirectly applies data augmentation on the training data by reversing the input clips in temporal order, helping the segmentation network to learn more general feature correspondences.

\subsection{Gradient correction}
\begin{figure*}
    \centering
    \includegraphics[width=\textwidth]{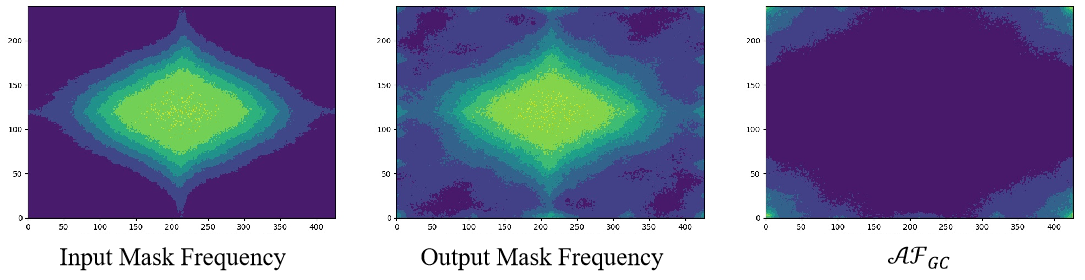}
    \caption{Frequency domain distribution of input mask, output mask and amplitude response of gradient correction module. \red{The results is obtained by averaging the response value on all frames of DAVIS17 validation set.}}
    \label{fig:freq}
\end{figure*}

\textbf{Cyclic refinement process.} After training with the cyclic loss as Equation~(\ref{eq:cycle-loss}), we can directly apply the offline model in the inference stage. However, inspired by the cyclic structure in training process, we can take the accurate initial reference mask as a measurement to evaluate the segmentation quality of current frame and proceed to refine the output results based on the evaluation results. In this way, we can explicitly reduce the effect of error propagation during inference time to keep our trained model from disturbances by natural or adversarial noises.

To achieve this goal, we design a gradient correction block to update segmentation results iteratively as illustrated in Figure~\ref{fig:framework}. Since only the initial mask $Y_1$ is available in inference stage, we apply the predicted mask $\widehat{Y}_t$ to infer the initial reference mask in the same manner as Equation~(\ref{eq:cycle}), and then evaluate the segmentation quality of $\widehat{Y}_t$ with the loss function in Equation~(\ref{eq:loss}). Intuitively, when more accurate prediction mask $\widehat{Y}_t$ are taken as reference, smaller reconstruction error for $Y_1$ will be yielded; therefore, during the gradient correction, our algorithm is focusing on minimizing the reconstruction error of the reference mask
\begin{equation}\label{eq:gc}
    \min_{\widehat{Y}_t}{\mathcal{L}_{rec}=\min_{\widehat{Y}_t}\mathcal{L}\left(\mathcal{S}_{\theta}\left(\{X_t\}, \{\widehat{Y}_{t}\}, X_1\right), Y_1\right)}
\end{equation}
Where the loss term $\mathcal{L}(\cdot, \cdot)$ adopts the same formulation as Equation~(\ref{eq:loss}) The gradient descent method is adopted to solve the reconstruction problem in Equation~(\ref{eq:gc}) so as to refine the mask $\widehat{Y}_t$. To be specific, we start from an output mask $\widehat{Y}^0_t=\widehat{Y}_t$, and then update the mask for $N$ iterations:
\begin{equation}\label{eq:gradient}
    \widehat{Y}_{t}^{l+1} = \widehat{Y}^{l}_{t} - \alpha \frac{\partial\mathcal{L}_{rec}}{\partial\widehat{Y}^{l}_{t}}
\end{equation}
where $\alpha$ is a predefined correction rate for mask update and $N$ is the iteration times. With this iterative refinement, we naturally extend the offline model to an online inference algorithm. However, the gradient correction approach can be time-consuming since it requires multiple times of network forward-backward pass. Due to this reason, we only apply gradient correction once per $K$ frames to achieve good performance-runtime trade-off.

\textbf{Interpretation in the frequency domain.} Empirically, with the gradient correction process in Equation~(\ref{eq:gradient}), the details of output mask can be better handled. This claim can be demonstrated from the aspect of frequency domain, where we find that the gradient correction module empirically acts a high-frequency amplifier to polish the output mask in fine-grained details. To analyze from the aspect of frequency, we take the gradient correction module as a black box system and calculate its frequency response by computing the averaged ratio between output and input amplitude-frequency characteristics,
\begin{equation}
    \mathcal{AF}_{GC} = \frac{1}{T}\sum_{t=1}^{T}\frac{\left|\mathcal{FFT}\left(\widehat{Y}^N_t\right)\right|}{\left|\mathcal{FFT}\left(\widehat{Y}^0_t\right)\right|}
\end{equation}
where $\mathcal{FFT}(\cdot)$ denotes 2D Fast Fourier transform. We visualized the 2D frequency-domain intensity of the input mask, output mask of gradient correction and the frequency-domain response $ \mathcal{AF}_{GC}$ in the form of amplitude-frequency characteristic contour in Figure~\ref{fig:freq}. We observe that compared with the input mask, the output mask manifests stronger intensity in high-frequency component (the four corner of the contour figure), the frequency response of gradient correction module also highlights and amplifies the part corresponding high-frequency harmonic. This observation provides the explanation of higher boundary accuracy improvement from the perspective of frequency since the high-frequency component usually supplements some detailed information of output masks.

\textbf{Anti-noise regularization.} However, intuitively, amplifying the high-frequency component will not only append details but also results in noise and artifacts around the object boundaries. To overcome such artifacts from gradient-correction, we augment the reconstruction error with a regularization term to suppress potential noise after refinement, denoted as
\begin{equation}\label{eq:regularize}
    {L}_{rec}=\mathcal{L}\left(\mathcal{S}_{\theta}\left(\{X_t\}, \{\widehat{Y}_{t}\}, X_1\right), Y_1\right) + \lambda \mathcal{L}_{smooth}\left(\widehat{Y}_t\right)
\end{equation}
where $ \mathcal{L}_{smooth}\left(\widehat{Y}_t\right)$ is a spatial smooth term to avoid exaggerating some spot-like area in output masks with $\lambda$ as its corresponding weight parameter.
\begin{equation}
    \mathcal{L}_{smooth}\left(\widehat{Y}_t\right) = \frac{1}{|\Omega|}\sum_{u \in \Omega}{\nabla_x \widehat{Y}_{t, u}^2 + \nabla_y \widehat{Y}_{t, u}^2}
\end{equation}
In instantiation, the Sobel operators $\nabla_x,\nabla_y$ are first applied on the mask $\widehat{Y}_t$ along the horizontal and vertical directions to calculate the spatial gradient, then areas with large spatial gradient norm are penalized. The augmented reconstruction objective can still be optimized according to the manner of Equation~(\ref{eq:gradient}).

\subsection{Cycle-ERF}
The cyclic mechanism with gradient update in Equation~(\ref{eq:gradient}) is not only helpful for the output mask refinement, but it also offers a new aspect of analyzing the region of interests of specific objects segmented by the pretrained network. In detail, we construct a reference set, $\mathcal{X}_l=\{X_l\}$ and $\mathcal{Y}_l=\{\mathbf{0}$\} as the guidance, where $\mathbf{0}$ denotes an empty mask of the same size as $X_l$ but is filled with zeros. We take these references to predict objects at the $t$-th frame $\widehat{Y}_t$. To this end, we can obtain the prediction loss $\mathcal{L}(\widehat{Y}_t, Y_t)$. To minimize this loss, we conduct the gradient correction process as in Equation~(\ref{eq:gradient}) to gradually update the empty mask for $M$ iterations. Finally, we take the ReLU function to preserve the positively activated areas of the objective mask as our final cycle-ERF representation, the resulting receptive field can be expressed as
\begin{equation}\label{eq:erf}
    \textbf{cycle-ERF}(Y_l) = ReLU\left(\widehat{Y}^{M}_{l}|_{\widehat{Y}^{0}_{l}=\mathbf{0}}\right)
\end{equation}

As we will show in our experiments, the cycle-ERF is capable of properly reflecting the support region of specific objects for the segmentation task. Through this analysis, the pretrained model can be shown to be particularly concentrated on certain objects in videos, making the learned segmentation model more interpretable.

\section{Experiments}

\subsection{Experiment setup}
\textbf{Datasets.} We train and evaluate our method on three widely used benchmarks for semi-supervised video object segmentation, DAVIS16~\cite{Perazzi2016}, DAVIS17~\cite{Pont-Tuset_arXiv_2017}, and Youtube-VOS~\cite{Xu_2018_S2S_ECCV}. DAVIS16 contains 50 videos in total, where 30 sequences are used for training and the others are taken as validation. In this benchmark, each video only covers a single reference mask. DAVIS17 is a multi-object extension of DAVIS16 and contains 120 video sequences in total with at most 10 objects in a video. The dataset is split into 60 sequences for training, 30 for validation, and the other 30 for test. The Youtube-VOS is larger in scale and contains more object categories. There are a total of 3,471 video sequences for training and 474 videos for validation in this dataset with at most 12 objects in a video, it also contains videos where objects appear from intermediate frames. Following the training procedure in~\cite{Oh_2019_ICCV,Voigtlaender_2019_CVPR}, we construct a hybrid training set by mixing the data from all training sequences. 

\textbf{Metrics.} For evaluation on DAVIS16, DAVIS17 validation, and test set, we adopt the metric following standard DAVIS evaluation protocol~\cite{Pont-Tuset_arXiv_2017}. The Jaccard overlap $\mathcal{J}$ is adopted to evaluate the mean IOU between predicted and groundtruth masks. The contour F-score $\mathcal{F}$ computes the F-measurement in terms of the contour based precision and recall rate. The final score is obtained from the average value of $\mathcal{J}$ and $\mathcal{F}$. The evaluation on Youtube-VOS follows the same protocol except that the two metrics are computed on seen and unseen objects respectively and averaged together.

\textbf{Baselines.} We take three recent methods as our base model for further analysis of our proposed cyclic mechanism in the following experiments.
\begin{itemize}
    \item Space Time Memory Network (STM)~\cite{Oh_2019_ICCV} is a widely used pipeline for fast and accurate semi-supervised video object segmentation, and the memory mechanism makes it flexible in adjusting reference sets $\mathcal{X}_t$ and $\mathcal{Y}_t$, which is suitable as comparison to our reference-based approach.
    \item Kernelized Memory Network (KMN)~\cite{KMN} is the kernelized version of STM network, where a Gaussian kernel is dynamically calculated before merging the knowledge from memory into current query feature.
    \item Global Context Memory Network (GCM)~\cite{GCVOS2020} extends the STM-like structure with a global temporal span, where the spatially global context of each reference frame is stored and dynamically updated in the memory.
    \item \red{Associate Objects with Transformer (AOT)~\cite{AOT2021} is a more advanced and efficient video object segmentation framework, where a long-short term cross attention structure is designed to help with parallel video object segmentation on multiple objects.}
\end{itemize}
\red{Since there is no public training code of~\cite{Oh_2019_ICCV,KMN,GCVOS2020}, we implement them by ourselves, and for ~\cite{AOT2021}, we directly implement our cycle mechanism from the public code from the author}. For STM, in order to adapt to the time-consuming gradient correction process, we take the lightweight design by reducing the intermediate memory feature dimension, \red{resizing the input resolution for inference to $240 \times 427$, which is $1/4$ of the size in original work~\cite{Oh_2019_ICCV} ($480 \times 854$), and then upsampling the output to original resolution by nearest interpolation}. For KMN, we replace the argmax operation in original paper with soft-argmax to make sure the network is end-to-end trainable. \red{For AOT, we modify the one-hot encode into soft labels to ensure the cycle loop is end-to-end differentiable.} For ease of representation, we denote the models trained with cyclic scheme with a ``-cycle'' suffix. It should be mentioned that the adopted baseline models from the original papers involved different external static data from various segmentation datasets~\cite{COCO,Everingham15}, resulting in unfair and inconsistent comparisons here. To enable fairer and more consistent comparison, for most of our analysis, we re-train our implemented models using only the training data in DAVIS and Youtube-VOS. \red{Nevertheless, for more comprehensive comparison, we still provide results of STM and AOT with the same setting as~\cite{Oh_2019_ICCV}, where the model is pretrained purely on COCO~\cite{COCO} and predict the object mask at the resolution of ($480 \times 854$), which is the most commonly adopted experimental setting.}

\textbf{Implementation details.} The training and inference procedures are deployed on an NVIDIA TITAN Xp GPU. Within an epoch, for each video sequence, we randomly sample 3 frames as the training samples -- the frame with the smallest timestamp is regarded as the initial reference frame. Similar to~\cite{Oh_2019_ICCV}, the maximum temporal interval of sampling increases by 5 every 20 training epochs. We set the hyperparameters as $\gamma=1.0$, $\lambda=0.75$, $N=10$, $K=5$, and $M=50$. During training, we adopt a bootstrapping strategy for the cross entropy loss, where only the top $40\%$ pixels with maximum training loss are taken into account. 
The ResNet series of models~\cite{He_2016_CVPR} pretrained on ImageNet~\cite{imagenet_cvpr09} are adopted as our backbone for baseline. The network is trained with a batch size of 4 for 240 epochs in total and is optimized by the Adam optimizer~\cite{adam2014method} of learning rate $10^{-5}$ and $\beta_1=0.9, \beta_2=0.999$. In both training and inference stages, the input frames are resized to the resolution of $240 \times 427$. The final output is upsampled to the original resolution by nearest interpolation. For simplicity, we directly use $X_{t}$ and $\widehat{Y}_{t}$ to construct the cyclic reference sets. For the case of multiple objects, we adopt the soft aggregation method in~\cite{Lin_2019_ICCV,Oh_2019_ICCV} to normalize the probability at each pixel of output masks.

\textbf{Data augmentation.} We apply common augmentation operations including random horizontal flipping, additive Gaussian noise and contrast enhancement. Additionally, we also adopt random crop strategy and fixed affine transformation (sheer, resize, rotation) to each training sample. We note that frames selected from the same video will share the same transformation parameters. When exploiting COCO to synthesize data, we follow~\cite{stm-coco} to take random affine and copy-paste trick to generate pseudo sequences.

\subsection{Main results}
\begin{table*}
\small
    \centering
    \begin{tabular}{c|ccc|ccc|c}\hline
        \multicolumn{7}{c}{DAVIS17 validation}\\\hline
        Method & Extra data & OL & GC & $\mathcal{J}$(\%) & $\mathcal{F}$(\%) & $\mathcal{J}\& \mathcal{F}$(\%) & FPS \\\hline
        RGMP~\cite{Oh_2018_CVPR} & \checkmark & & & 64.8 & 68.6 & 66.7 & 3.6 \\
        DMM-Net~\cite{Zeng_2019_ICCV} & \checkmark & & & 68.1 & 73.3 & 70.7 & - \\
        AGSS-VOS~\cite{Lin_2019_ICCV} & \checkmark & & & 63.4 & 69.8 & 66.6 & 10 \\
        AGSS-VOS~\cite{Lin_2019_ICCV} & \checkmark & & \checkmark & 64.0 & 70.6 & 67.3 & 2.2 \\
        FEELVOS~\cite{Voigtlaender_2019_CVPR} & \checkmark & & & 69.1 & 74.0 & 71.5 & 2 \\
        FRTM~\cite{Robinson_2020_CVPR} & & & & - & - & 70.2 & 41.3 \\
        \hline
        OnAVOS~\cite{voigtlaender17BMVC} & \checkmark & \checkmark & & 61.0 & 66.1 & 63.6 & 0.04 \\
        PReMVOS~\cite{luiten2018premvos} & \checkmark & \checkmark & & {73.9} & {81.7} & {77.8} & 0.03 \\\hline
        STM-ResNet50~\cite{Oh_2019_ICCV}$^\dagger$ & \checkmark & & & 79.7 & 84.4 & 82.0 & 7.9 \\
        STM-ResNet50-cycle (Ours)$^{\dagger}$ & \checkmark & & \checkmark & \textbf{80.4} & \textbf{85.2} & \textbf{82.8} & 2.5 \\ \hline
        STM-ResNet18-cycle (Ours) & & & & 64.7 & 69.9 & 67.3 & \textbf{55.3} \\
        STM-ResNet18-cycle (Ours) & & & \checkmark & 65.3 & 70.8 & 68.1 & 13.4 \\
        STM-ResNet50-cycle (Ours) & & & & 68.7 & 74.7 & 71.7 & 38 \\
        STM-ResNet50-cycle (Ours) & & & \checkmark & 69.8 & 75.9 & 72.9 & 9.3 \\\hline\hline
        \multicolumn{7}{c}{DAVIS17 test-dev} \\\hline
        Method & Extra data & OL & GC & $\mathcal{J}$(\%) & $\mathcal{F}$(\%) & $\mathcal{J}\& \mathcal{F}$(\%) & FPS \\\hline
        RVOS~\cite{Ventura_2019_CVPR} & & & & 48.0 & 52.6 & 50.3 & 22.7 \\
        RGMP~\cite{Oh_2018_CVPR} & \checkmark & & & 51.3 & 54.4 & 52.8 & 2.4 \\
        AGSS-VOS~\cite{Lin_2019_ICCV} & \checkmark & & & 54.8 & 59.7 & 57.2 & 10 \\
        FEELVOS~\cite{Voigtlaender_2019_CVPR} & \checkmark & & & 55.2 & 60.5 & 57.8 & 1.8 \\ \hline
        OnAVOS~\cite{voigtlaender17BMVC} & \checkmark & \checkmark & & 53.4 & 59.6 & 56.9 & 0.03 \\
        PReMVOS~\cite{luiten2018premvos} & \checkmark & \checkmark & & 67.5 & 75.7 & 71.6 & 0.02 \\\hline
        STM-ResNet50~\cite{Oh_2019_ICCV}$^\dagger$ & \checkmark & & & 68.0 & 74.1 & 71.0 & 14.8 \\
        STM-ResNet50-cycle (Ours)$^{\dagger}$ & \checkmark & & \checkmark & \textbf{70.6} & \textbf{76.4} & \textbf{73.5} & 4.1 \\ \hline
        STM-ResNet18-cycle (Ours) & & & & 53.2 & 58.4 & 55.8 & \textbf{44.7} \\
        STM-ResNet18-cycle (Ours) & & & \checkmark & 53.7 & 60.5 & 57.2 & 10.7\\
        STM-ResNet50-cycle (Ours) & & & & 55.1 & 60.5 & 57.8 & 31 \\
        STM-ResNet50-cycle (Ours) & & & \checkmark & 55.4 & 62.8 & 59.1 & 6.9 \\\hline\hline
        \multicolumn{7}{c}{DAVIS16 validation} \\\hline
        Method & Extra data & OL & GC & $\mathcal{J}$(\%) & $\mathcal{F}$(\%) & $\mathcal{J}\& \mathcal{F}$(\%) & FPS \\\hline
        Lucid Dreaming~\cite{lucid_dreaming} & & & & 83.9 & 82.0 & 83.0 & - \\
        RGMP~\cite{Oh_2018_CVPR} & \checkmark & & & 81.5 & 82.0 & 81.8 & 7.8 \\
        AGAME~\cite{agame} & \checkmark & & & 81.5 & 82.2 & 81.9 & 14.3 \\
         FEELVOS~\cite{Voigtlaender_2019_CVPR} & \checkmark & & & 81.1 & 82.2 & 81.7 & - \\ 
         FRTM~\cite{Robinson_2020_CVPR} & & & & - & - & 78.5 & 41.3 \\\hline
         OnAVOS~\cite{voigtlaender17BMVC} & \checkmark & \checkmark & & 86.1 & 84.9 & 85.5 & 0.08 \\
        PReMVOS~\cite{luiten2018premvos} & \checkmark & \checkmark & & 84.9 & 88.6 & 86.8 & 0.02 \\\hline
        STM-ResNet50~\cite{Oh_2019_ICCV}$^\dagger$ & \checkmark & & & 88.9 & 88.9 & 88.9 & 7.4 \\
        STM-ResNet50-cycle (Ours)$^{\dagger}$ & \checkmark & & \checkmark & \textbf{89.2} & \textbf{90.4} & \textbf{89.8} & 2.0 \\ \hline
        STM-ResNet18-cycle (Ours) & & & & 80.4 & 80.3 & 80.4 & \textbf{64.5} \\
        STM-ResNet18-cycle (Ours) & & & \checkmark & 81.3 & 81.1 & 81.2 & 17.7 \\
        STM-ResNet50-cycle (Ours) & & & & 84.1 & 83.7 & 83.9 & 38.5 \\
        STM-ResNet50-cycle (Ours) & & & \checkmark & 84.1 & 83.8 & 84.0 & 11.5 \\\hline
    \end{tabular}
    \caption{Comparison with state-of-the-art method on DAVIS16 and DAVIS17 set. ``Extra data'' indicates the method is pretrained with extra data with mask annotations.``-'' indicates unavailable results. ``OL'' denotes online learning or update process. ``GC'' is short for gradient correction. \red{``$\dagger$'' denotes the pretraining is implemented from~\cite{stm-coco} with larger input size $(480 \times 854)$ for inference.}}\label{tab:davis} 
\end{table*}

\begin{table*}
\small
    \centering
    \begin{tabular}{c|ccc|ccccc|c}\hline
        Method & Extra data & OL & GC & $\mathcal{J}_{\mathcal{S}}$(\%) & $\mathcal{J}_{\mathcal{U}}$(\%) & $\mathcal{F}_{\mathcal{S}}$(\%) & $\mathcal{F}_{\mathcal{U}}$(\%) & $\mathcal{G}$(\%) & FPS\\\hline
        RVOS~\cite{Ventura_2019_CVPR} & & & & 63.6 & 45.5 & 67.2 & 51.0 & 56.8 & 24 \\
        S2S~\cite{Xu_2018_S2S_ECCV} & & & & 66.7 & 48.2 & 65.5 & 50.3 & 57.6 & 6 \\
        FRTM~\cite{Robinson_2020_CVPR} & & & & 68.6 & 58.4 & 71.3 & 64.5 & 65.7 & - \\
        TVOS~\cite{zhang2020a} & & & & 67.1 & 63.0 & 69.4 & 71.6 & 67.8 & - \\
        RGMP~\cite{Oh_2018_CVPR} & \checkmark & & & 59.5 & - & 45.2 & - & 53.8 & 7 \\
        DMM-Net~\cite{Zeng_2019_ICCV} & \checkmark & & & 58.3 & 41.6 & 60.7 & 46.3 & 51.7 & 12 \\
        AGSS-VOS~\cite{Lin_2019_ICCV} & \checkmark & & & 71.3 & {65.5} & 75.2 & {73.1} & {71.3} & 12.5 \\
        \hline
        S2S~\cite{Xu_2018_S2S_ECCV} & & \checkmark & & 71.0 & 55.5 & 70.0 & 61.2 & 64.4 & 0.06 \\
        OSVOS~\cite{Cae_OVOS_17} & \checkmark & \checkmark & & 59.8 & 54.2 & 60.5 & 60.7 & 58.8 & - \\
        MaskTrack~\cite{Perazzi_2017_CVPR} & \checkmark & \checkmark & & 59.9 & 45.0 & 59.5 & 47.9 & 53.1 & 0.05 \\
        OnAVOS~\cite{voigtlaender17BMVC} & \checkmark & \checkmark & & 60.1 & 46.6 & 62.7 & 51.4 & 55.2 & 0.05 \\
        DMM-Net~\cite{Zeng_2019_ICCV} & \checkmark & \checkmark & & 60.3 & 50.6 & 63.5 & 57.4 & 58.0 & - \\\hline
        STM-ResNet50~\cite{Oh_2019_ICCV}$^{\dagger}$ & \checkmark & & & 76.1 & 70.8 & 79.6 & 77.5 & 76.0 & 3.9 \\  
        STM-ResNet50-cycle (Ours)$^{\dagger}$ & \checkmark & & \checkmark & \textbf{77.8} & \textbf{73.3} & \textbf{81.5} & \textbf{80.1} & \textbf{78.2} & 0.9 \\
        \hline
        STM-ResNet18-cycle (Ours) & & & & 69.2 & 56.2 & 72.5 & 65.0 & 65.7 & \textbf{63} \\
        STM-ResNet18-cycle (Ours) & & & \checkmark & {70.4} & 58.2 & 73.9 & 67.2 & 67.5 & 15.2 \\
        STM-ResNet50-cycle (Ours) & & & & 71.7 & 61.4 & 75.8 & 70.4 & 69.9 & 43 \\
        STM-ResNet50-cycle (Ours) & & & \checkmark & {72.6} & 63.0 & 76.7 & 72.3 & 71.2 & 13.8 \\\hline
    \end{tabular}
    \caption{Comparison with state-of-the-art method on Youtube-VOS validation set. The subscript $\mathcal{S}$ and $\mathcal{U}$ denote the seen and unseen categories.  $\mathcal{G}$ is the global mean.  ``-'' indicates unavailable results.``OL'' denotes online learning or update process. ``GC'' is short for gradient correction. \red{``$\dagger$'' denotes the pretraining is implemented from~\cite{stm-coco} with larger input size $(480 \times 854)$ for inference.}}
    \label{tab:youtube}
    \vspace{-5mm}
\end{table*}

In this section, we first report the 
comparison results between our cyclic model and other methods, where our full model is measured with configuration of different backbone and cyclic schemes.

\textbf{DAVIS.} The evaluation results on DAVIS16 validation set, DAVIS17 validation and test-dev set are reported in Table~\ref{tab:davis}. From this table, we observe that our model trained with cyclic loss outperforms most of the offline methods and even performs better than the method with online learning~\cite{voigtlaender17BMVC} on DAVIS17 benchmark. When combined with the online gradient correction process, our method gets further improvement. \red{It should also be noticeable that although standard STM-cycle do not require additional training data other than DAVIS and Youtube-VOS, it outperforms some other state-of-the-art pipelines highly dependent on additional training data~\cite{Lin_2019_ICCV,Oh_2018_CVPR,Voigtlaender_2019_CVPR}.} In terms of the runtime speed, although gradient correction increases the computation cost, our method still runs at a speed comparable to other offline methods~\cite{Lin_2019_ICCV} due to our efficient implementation. When we replace the backbone network from ResNet50 to more lightweight ResNet18, our trained model can run faster with still competitive performance. Although there is a performance gap between our approach that is trained from scratch and the state-of-the-art online learning method, our method is far more efficient and it does not requires collecting extra data from instance segmentation tasks as training samples. \blue{It is also noticeable that when adding COCO into training set, cyclic version of STM can achieve state-of-the-art performance on all benchmarks of DAVIS, obtaining consistent improvement over original STM~\cite{Oh_2019_ICCV}, which shares the same backbone but requires more data besides COCO}. 

Furthermore, we also try to combine the gradient correction process with an existing open source model of AGSS-VOS~\cite{luiten2018premvos}\footnote{\url{https://github.com/Jia-Research-Lab/AGSS-VOS}} to test inference performance gain on state-of-the-art method with purely gradient correction. This appears to bring improvement on the overall segmentation quality, demonstrating that cyclic consistency is helpful even in different segmentation pipelines.

\textbf{Youtube-VOS.} The evaluation results on Youtube-VOS validation set are reported in Table~\ref{tab:youtube}. On this benchmark, our model also outperforms some offline methods and their online learning counterparts~\cite{Xu_2018_S2S_ECCV,Zeng_2019_ICCV}. It is also noticeable that compared with the performance on seen objects, the one on unseen objects has improved more using our gradient correction strategy. Further, we observe that with ResNet-50 backbone and gradient correction technique, our final pipeline can achieve similar segmentation accuracy and runtime speed to some state-of-the-art methods ~\cite{Lin_2019_ICCV} on Youtube-VOS even without extra data for pre-training. Finally, as consistent with the results on DAVIS series, when combined with COCO pretraining, our STM-cycle model can achieves better performance than other state-of-the-art models with pretrained~\cite{Oh_2019_ICCV,Lin_2019_ICCV,Oh_2018_CVPR}

\subsection{Ablation study}
\begin{table*}
    \centering
    {
    \begin{tabular}{cc|cc|c|cc|c}\hline
        \multicolumn{2}{c|}{datasets} & \multicolumn{3}{|c|}{DAVIS17} & \multicolumn{3}{|c}{DAVIS16} \\\hline 
       $\mathcal{X}_{t-1}$ & $\mathcal{Y}_{t-1}$ & baseline & +cycle & $\Delta$ & baseline & +cycle & $\Delta$\\\hline
        $\{X_1\}$ & $\{Y_1\}$ & 65.2 & 67.6 & +2.4 & 81.2 & 81.2 & +0.0 \\
        $\{X_{t-1}\}$ & $\{\widehat{Y}_{t-1}\}$ & 56.8 & 61.2 & \textbf{+4.4} & 75.3 & 77.2 & \textbf{+1.9}\\
        $\{X_1, X_{t-1}\}$ & $\{Y_1, \widehat{Y}_{t-1}\}$ & 67.3 & 69.2 & +1.9 & 82.3 & 83.8 & +1.5\\ \textbf{MEM} & \textbf{MEM} & 69.7 & 71.7 & +2.0 & 82.3 & 83.9 & +1.6 \\\hline
    \end{tabular}
    }
    \caption{Experiments on improvement of $\mathcal{J}\&\mathcal{F}$ score with different reference set configuration.}\label{tab:config}
\end{table*}
\begin{table}
    \centering
    {
    \begin{tabular}{c|c|ccc}\hline
         & datasets & $\mathcal{J}$(\%) & $\mathcal{F}$(\%) & $\mathcal{J} \& djk\mathcal{F}$(\%) \\\hline
        baseline & \multirow{3}{*}{DAVIS17} & 67.6 & 71.7 & 69.7 \\
        + cyclic & & 68.7 & 74.7 & 71.7 \\
        + GC & & 69.2 & 74.3 & 71.8 \\
        + both & & \textbf{69.8} & \textbf{75.9} & \textbf{72.9} \\\hline
         baseline & \multirow{3}{*}{DAVIS16} & 82.8 & 81.7 & 82.3 \\
        + cyclic & & 84.1 & 83.7 & 83.9 \\
        + GC & & 83.3 & 82.3 & 82.8 \\
        + both & & \textbf{84.1} & \textbf{83.8} & \textbf{84.0} \\\hline
    \end{tabular}
    }
    \caption{Ablation study on the effectiveness of different component. ``GC'' is short for gradient correction.}
    \label{tab:compnent}
\end{table}

In this section, we conduct a series of experiments to analyze the cyclic property of our trained models, with all the results evaluated on the DAVIS16 and DAVIS17 validation set and ResNet50 as backbone network.

\subsubsection{Effectiveness of each cyclic component.} 
We first demonstrate the effectiveness of cyclic training and gradient correction in Table~\ref{tab:compnent}, where the baseline method~\cite{Oh_2019_ICCV} based on STM network is re-implemented and retrained.
From this table, both components are shown to be helpful in boosting the performance on both DAVIS16 and DAVIS17 validation sets. In particular, the incorporated cycle mechanism improves the contour score $\mathcal{F}$ more than the overlap score $\mathcal{J}$, signifying that the proposed scheme is likely to be more useful for fine-grained mask prediction. 

\subsubsection{Improvement with different reference sets.} 
Due to the flexibility of our baseline method in configuring its reference sets during inference, we tested how our cyclic training strategy would impact VOS performance using different reference sets on our STM baseline. We conduct the test under four types of configuration: (1) Only the initial reference mask and its frame are utilized for predicting other frames. (2) Only the prediction of the last frame $\widehat{Y}_{t-1}$ and the last frame are used. (3) Both the initial reference and last frame prediction are utilized, which is the most common configuration in other state-of-the-art works. (4) The external memory strategy (denoted as \textbf{MEM}) in~\cite{Oh_2019_ICCV} is used where the reference set is dynamically updated by appending new prediction and frames at a specific frequency of $5$Hz. In the results reported in Table~\ref{tab:config}, we observe that the cyclic training is helpful under all configurations. It is also interesting to see that our scheme achieves the maximum improvement (+4.6 $\mathcal{J}\&\mathcal{F}$ on DAVIS17 and +1.9 $\mathcal{J}\&\mathcal{F}$ on DAVIS16) with the configuration $\mathcal{X}_{t-1}=\{X_{t-1}\}, \mathcal{Y}_{t-1}=\{\widehat{Y}_{t-1}\}$, since this case is the most vulnerable to accumulative error propagation and hence, proper training with cyclic loss term can effectively relieve such a problem.

\begin{table*}
    \centering
    \begin{tabular}{c|c|ccc|c|ccc}\hline
        {datasets} & \multicolumn{4}{|c|}{DAVIS17} & \multicolumn{4}{|c}{DAVIS16} \\\hline
        Noise & +GC & $\mathcal{J}$(\%) & $\mathcal{F}$(\%) & $\mathcal{J}\&\mathcal{F}$(\%) & +GC & $\mathcal{J}$(\%) & $\mathcal{F}$(\%) & $\mathcal{J}\&\mathcal{F}$(\%)\\\hline
        \multirow{2}{*}{Low-quality} & & 65.9 & 72.2 & 69.1 & &  65.9 & 72.2 & 69.1\\
        & \checkmark & \textbf{66.9} & \textbf{73.3} & \textbf{70.1} & \checkmark & \textbf{66.9} & \textbf{73.3} & \textbf{70.1}\\\hline
        \multirow{2}{*}{Box-template} & & 63.0 & 66.0 & 64.5 & & 63.0 & 66.0 & 64.5\\
        & \checkmark & \textbf{68.7} & \textbf{74.9} & \textbf{71.8} & \checkmark & \textbf{68.7} & \textbf{74.9} & \textbf{71.8}\\\hline
    \end{tabular}
    \caption{Results of models affected by natural noisy masks on DAVIS17 and DAVIS16 validation set.}
    \label{tab:low-quality}
\end{table*}
\begin{table*}
    \centering
    \begin{tabular}{c|c|ccc|c|ccc}\hline
        {setting} & \multicolumn{4}{|c|}{White Box} & \multicolumn{4}{|c}{Black Box} \\\hline
        Noise & +GC & $\mathcal{J}$(\%) & $\mathcal{F}$(\%) & $\mathcal{J}\&\mathcal{F}$(\%) & +GC & $\mathcal{J}$(\%) & $\mathcal{F}$(\%) & $\mathcal{J}\&\mathcal{F}$(\%)\\\hline
        \multirow{2}{*}{FGSM~\cite{goodfellow2014explaining}} & & 43.2 & 48.8 & 46.0 & &  39.0 & 46.4 & 42.7\\
        & \checkmark & \textbf{52.7} & \textbf{59.4} & \textbf{56.0} & \checkmark & \textbf{51.0} & \textbf{59.7} & \textbf{55.4}\\\hline
        \multirow{2}{*}{MI-FGSM~\cite{dong2018boosting}} & & 38.6 & 44.8 & 41.7 & & 30.9 & 36.8 & 33.9\\
        & \checkmark & \textbf{50.6} & \textbf{58.0} & \textbf{54.3} & \checkmark & \textbf{47.7} & \textbf{55.6} & \textbf{51.7}\\\hline
    \end{tabular}
    \caption{Results of models affected by adversarial noise on DAVIS17 validation set.}
    \label{tab:adv}
\end{table*}
\begin{table*}
    \centering
    \begin{tabular}{c|cc|ccc|ccc}\hline
        {datasets} & \multicolumn{2}{|c|}{setting} & \multicolumn{3}{|c|}{DAVIS17} & \multicolumn{3}{|c}{DAVIS16} \\\hline
        base model & +cycle & +GC & $\mathcal{J}$(\%) & $\mathcal{F}$(\%) & $\mathcal{J}\&\mathcal{F}$(\%) & $\mathcal{J}$(\%) & $\mathcal{F}$(\%) & $\mathcal{J}\&\mathcal{F}$(\%)\\\hline
        \multirow{3}{*}{STM~\cite{Oh_2019_ICCV}} & & & 67.6 & 71.7 & 69.7 & 82.8 & 81.8 & 82.3 \\
        & \checkmark & & 68.7 & 74.7 & 71.7 & 84.1 & 83.5 & 83.8 \\
        & \checkmark & \checkmark & \textbf{69.8} & \textbf{75.9} & \textbf{72.9} & \textbf{84.1} & \textbf{83.8} & \textbf{84.0}\\\hline
        
        \multirow{3}{*}{KMN~\cite{KMN}} & & & 67.5 & 72.3 & 69.9 & 81.5 & 80.3 & 80.9\\
        & \checkmark & & 67.8 & 73.5 & 70.7 & 82.6 & 83.4 & 83.0 \\
        & \checkmark & \checkmark & \textbf{69.1} & \textbf{75.3} & \textbf{72.2} & \textbf{82.8} & \textbf{83.7} & \textbf{83.3}\\\hline
        
        \multirow{3}{*}{GCM~\cite{GCVOS2020}} & & & 66.7 & 72.9 & 69.8 & 81.1 & 81.1 & 81.1\\
        & \checkmark & & 67.5 & 73.3 & 70.4 & 83.3 & 83.2 & 83.3 \\
        & \checkmark & \checkmark & \textbf{67.6} & \textbf{73.7} & \textbf{70.7} & \textbf{83.6} & \textbf{83.6} & \textbf{83.6}\\\hline
        
        \multirow{3}{*}{AOT-T~\cite{AOT2021}} & & & 76.5 & 81.9 & 79.2 & 86.5 & 88.4 & 87.5\\
        & \checkmark & & 77.7 & 82.7 & 80.2 & 86.5 & 88.5 & 87.5 \\
        & \checkmark & \checkmark & \textbf{77.9} & \textbf{83.6} & \textbf{80.8} & \textbf{86.6} & \textbf{88.5} & \textbf{87.6}\\\hline
        
    \end{tabular}
    \caption{Results of cyclic mechanism with different baseline models on DAVIS17 and DAVIS16 validation sets. The baseline results of STM, KMN and GCM are from our own implementation. The results of AOT-T are obtained from the official public code.}
    \label{tab:extension}
\end{table*}

\begin{figure*}[t]
    \begin{minipage}{0.48\linewidth}
        \includegraphics[width=\textwidth]{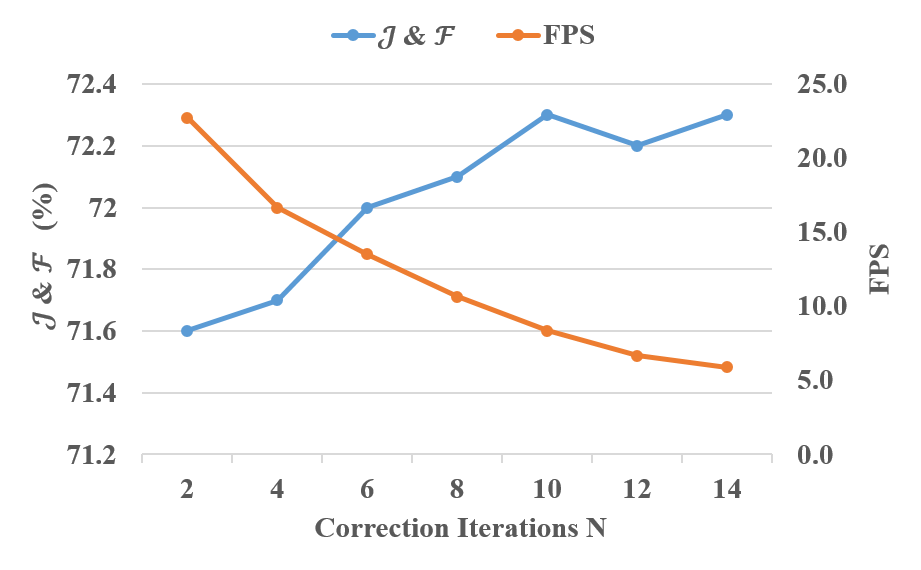}
    \caption{\red{Performance-runtime trade-off with different iteration size $N$ on DAVIS17 validation.}}
    \label{fig:runtime}
    \end{minipage}
    \noindent
    \begin{minipage}{0.48\linewidth}
        \includegraphics[width=\textwidth]{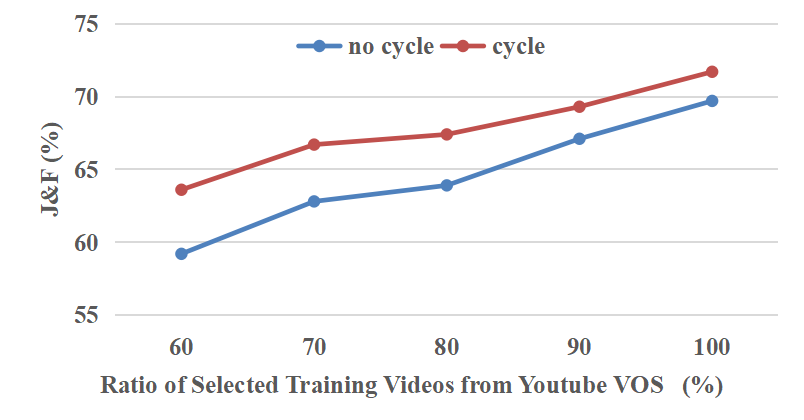}
    \caption{\red{Results on DAVIS17 validation set with different amount of training data in Youtube-VOS for STM-cycle and baseline STM model.}}
    \label{fig:data_size}
    \end{minipage}
\end{figure*}

\subsubsection{Sensitivity analysis.} 
\begin{figure}
    \includegraphics[width=0.48\textwidth]{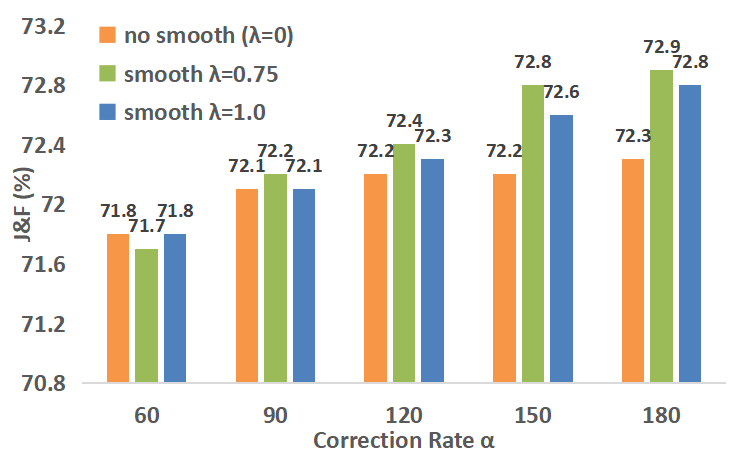}
\caption{\red{Performance with different correction rate and smooth term for STM-cycle. The model is trained on the hybrid dataset of DAVIS2017 and Youtube-VOS. And the evaluation results are reported on DAVIS2017 validation set.}}
\label{fig:alpha}
\end{figure}

Next, we evaluate how the hyperparameters in our algorithm affect the final results. In Figure~\ref{fig:runtime}, we show the performance-runtime trade-off w.r.t. the correction iteration time $N$. We find that the $\mathcal{J}\&\mathcal{F}$ score saturates 
when $N$ approaches 10; above which, the score improvement is somewhat marginal but at the expense of decreasing efficiency. \red{Considering the trade-off between runtime and performance, we take $N=10$ as the empirically optimal iteration number for gradient correction. Additionally, we also analyze the impact of correction rate $\alpha$ as shown in Figure~\ref{fig:alpha}. From this figure, we find the larger strength of correction usually results in better performance, but the overall performance variation is not sensitive to the change of correction rate $\alpha$, reflecting that our update scheme is robust and can accommodate variations to this parameter well, consequently, we set correction rate $\alpha=180$ since the overall gain becomes saturated under this configuration.}

\subsubsection{Effect of Anti-noise regularization}
\begin{figure}
    \centering
    \includegraphics[width=0.48\textwidth]{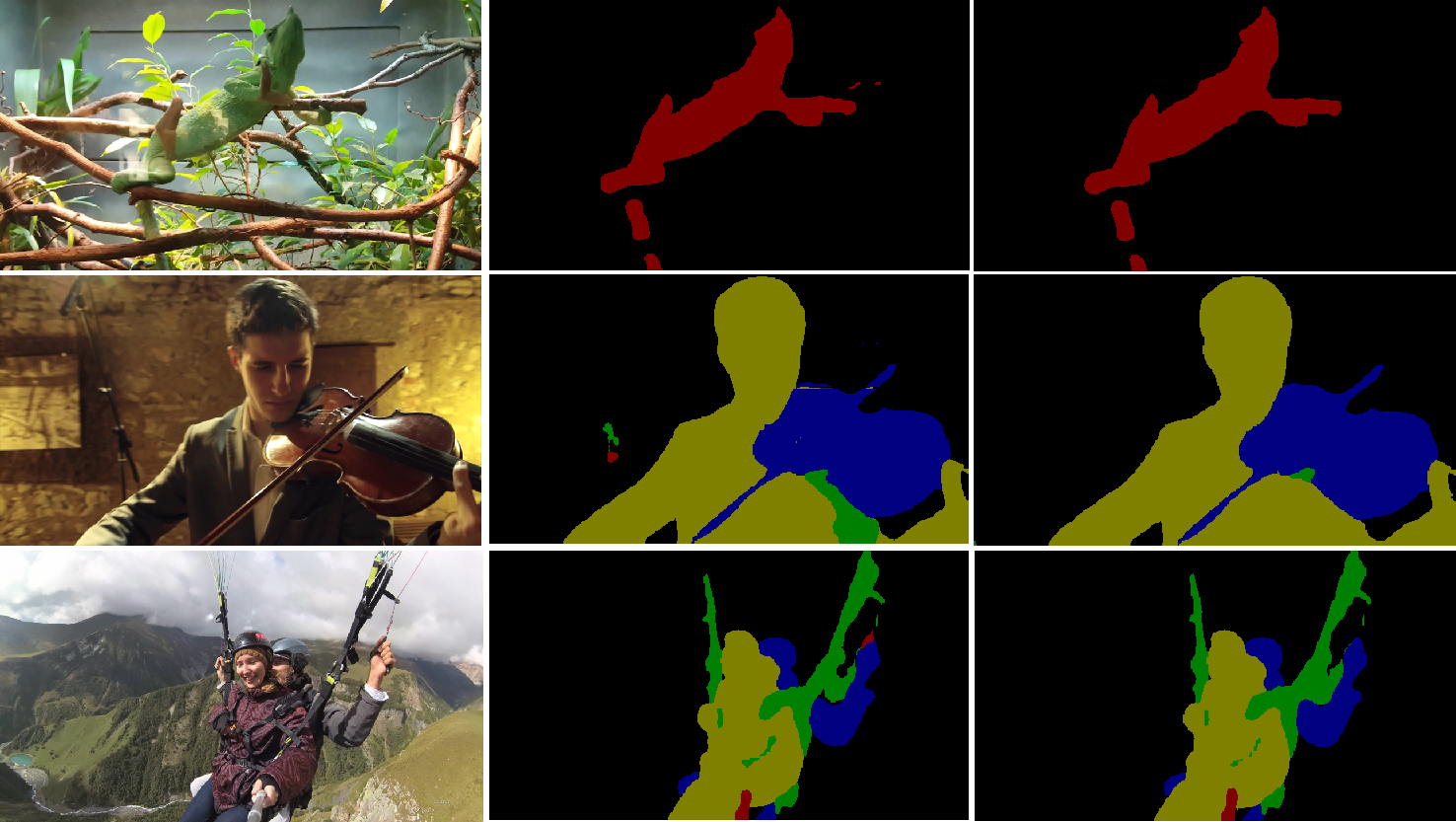}
    \caption{\red{Qualitative comparison between predicted results from gradient correction module with and without anti-noise regularization on DAVIS17 test-dev. Left column: The original query frame. Medium column: Predicted mask from gradient correction module without anti-noise regularization. Right column: Predicted mask from gradient correction module with anti-noise regularization.}}
    \label{fig:reg}
\end{figure}

\red{Together with the analysis of correction rate $\alpha$, we also investigate the effect of proposed anti-noise regularization term under different correction rate and loss weight $\lambda$. We find that the smooth term in Equation~(\ref{eq:gc}) can bring stable gains to overall segmentation quality except when the correction rate is small. When $\alpha$ is larger, the gain appears to be more obvious and finally converges to a stable level of improvement. We think this is because larger correction rates produces more precise segmentation but also tends to result in more severe background artifacts, which can be alleviated by the smooth constraint. Finally, we set $\lambda=0.75$ since we find the overall gain is not sensitive within proper interval of this loss weight.} 

\red{In Figure~\ref{fig:reg}, we further qualitatively analysis the impact of the regularization term during gradient correction. By comparison between the output masks, we can find we equipped with the smooth regularization, our gradient correction can suppress some subtle noise blocks in the background.}

\red{
\subsubsection{Results with Different Amount of Training Data}
As discussed in Section~\ref{sec:cycle-loss}, the cycle-consistency loss in our method implicitly applies data augmentation to train more generalized VOS model, thus should be more robust to scarcity of training data, especially for visual tasks that requires large amounts of synthesized data~\cite{COCO} for pretraining like VOS. To demonstrate this discussion, we specifically design experiments to test the performance when the training data is gradually decreased. In detail, we start from our standard benchmark with hybrid training set of DAVIS and Youtube-VOS ($\%100$ selected), then we gradually decrease the available training clips in Youtube-VOS (from $\%100$ to $\%60$) and evaluate the performance of trained model. The results are depicted in Figure~\ref{fig:data_size}. \blue{We can observe that when combined cyclic consistency loss, the performance of trained STM model degrades slower than the counterpart without cycle-consistency but with the same backbone, demonstrating that the cycle consistency can alleviate the problem of data scarcity to some extent.}}

\subsubsection{Robustness to noise perturbations} 
In addition to accumulative predicted error, video object segmentation systems are also prone to noise perturbations on reference templates. In this section, we further investigate how the gradient correction process mitigates the effects from such noisy interference. To do this, we utilize STM network with the \textbf{MEM} strategy as~\cite{Oh_2019_ICCV} by dynamically appending a predicted mask and its frame into the reference set. However, in this case, the predicted masks to be appended are manually replaced by a noisy version. Technically, we try to perturb the inference process with four types of noises, of which two are regarded as natural noises and the other two are adversarial-type noises.
\begin{itemize}
    \item \textbf{Low-quality}. We replace the predicted mask $\widehat{Y}_t$ from baseline model with lower segmentation quality on the same frame. 
    \item \textbf{Box-template}. We replace the predicted mask $\widehat{Y}_t$ with a coarse level groundtruth mask where all pixels in the bounding box of objects are set to be $1$.
    \item \textbf{FGSM}~\cite{goodfellow2014explaining}. We take the classical adversarial attack method to generate interference noise on intermediate reference masks, where the loss term in Equation~(\ref{eq:loss}) on all frames in a video is maximized under a given pixel-wise changing constraint $\epsilon \leq 20$.
    \item \textbf{MI-FGSM}~\cite{dong2018boosting}. We further added a harsher adversarial attack method to test the robustness of our model. The MI-FSGM method generates noise towards the same objective and constraint as FSGM but updates the reference mask according to an iterative manner with momentum.
\end{itemize}

 For the noise types generated by adversarial attack -- FGSM~\cite{goodfellow2014explaining} and MI-FGSM~\cite{dong2018boosting}, we follow the common protocol in adversarial attack and defence by testing the results under both black box and white box settings. In white box setting, we take the trained STM-ResNet50-cycle model to generate noise and the attack itself, while in black box setting, we leverage the trained STM-ResNet18-cycle model to generate noise and attack the one with the vanilla ResNet50 backbone. For each scheme, we conduct another experiment with gradient correction on the noisy masks before appending to the memory as the control group. The results are reported in Table~\ref{tab:low-quality} and Table~\ref{tab:adv}. From Table~\ref{tab:low-quality}, we see the gradient correction is helpful for both low-quality and box-template reference conditions. The improvement is much more obvious for the case of box-template, which indicates that the impact of gradient correction is greater when the intermediate reference mask is coarser but properly covers the object area. Meanwhile, from Table~\ref{tab:adv}, we could see that although both attack methods degrade the segmentation performance to a large extent, but by properly utilizing the gradient correction, we can alleviate the effect from adversarial attacks.
 
\begin{figure*}
    \centering
    \includegraphics[width=\textwidth]{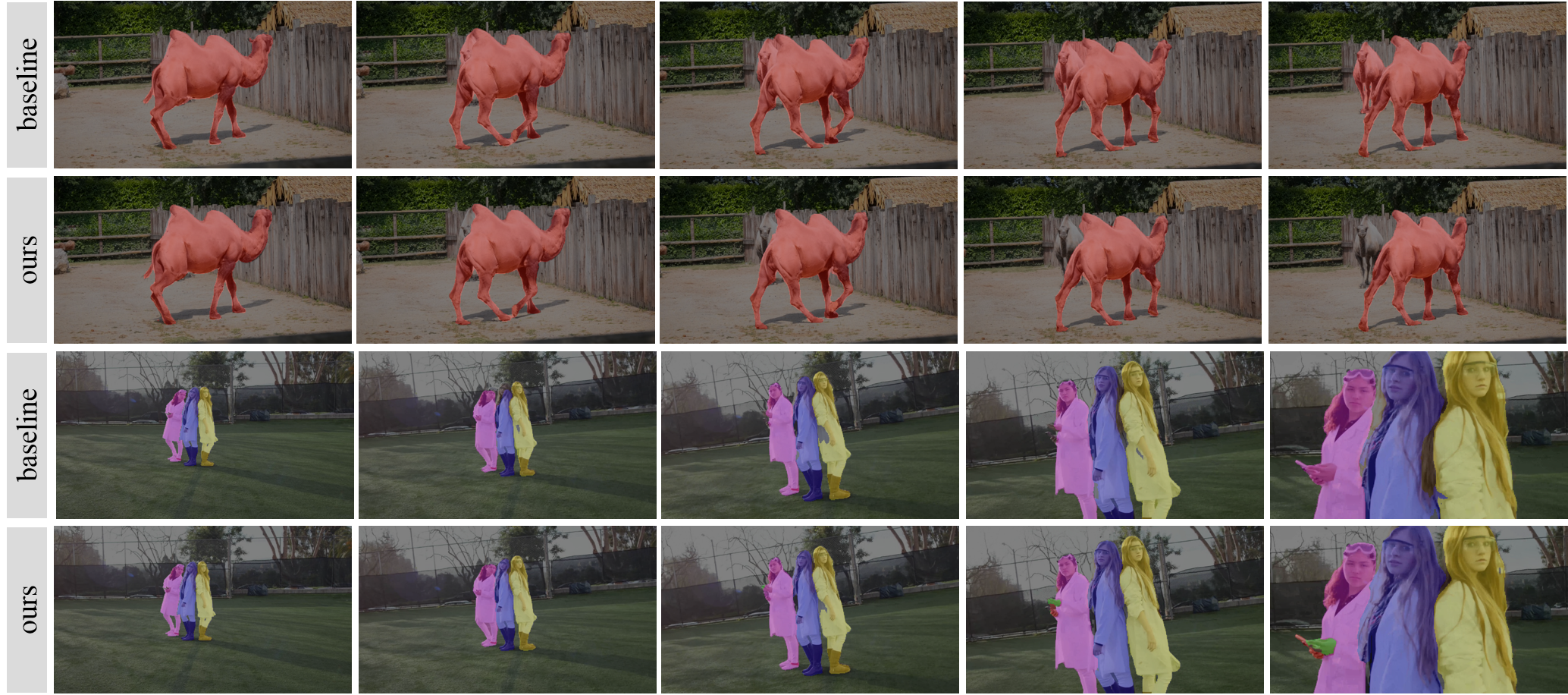}
    \caption{Qualitative results shows the improvement of cyclic training over the baseline in DAVIS17 validation.}
    \label{fig:vis}
\end{figure*}
\begin{figure*}
    \centering
    \includegraphics[width=0.96\textwidth]{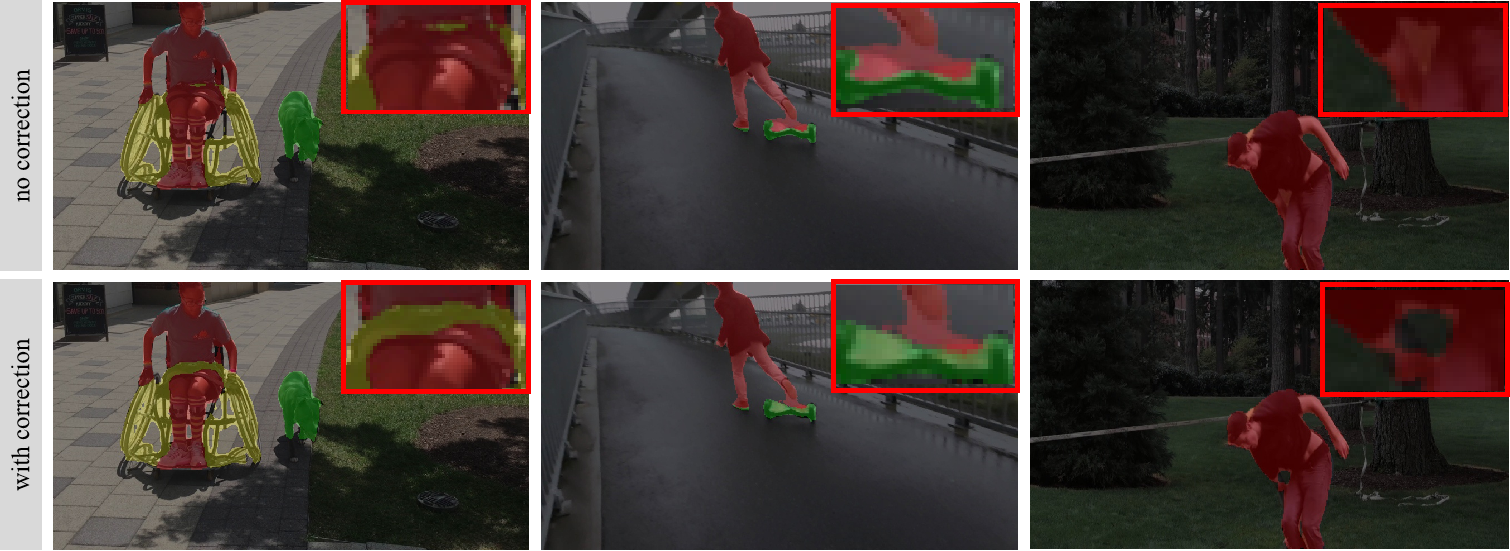}
    \caption{The visual effect of gradient correction module on DAVIS17 test-dev set}
    \label{fig:correction}
\end{figure*}
\begin{figure*}
    \centering
    \includegraphics[width=0.96\textwidth]{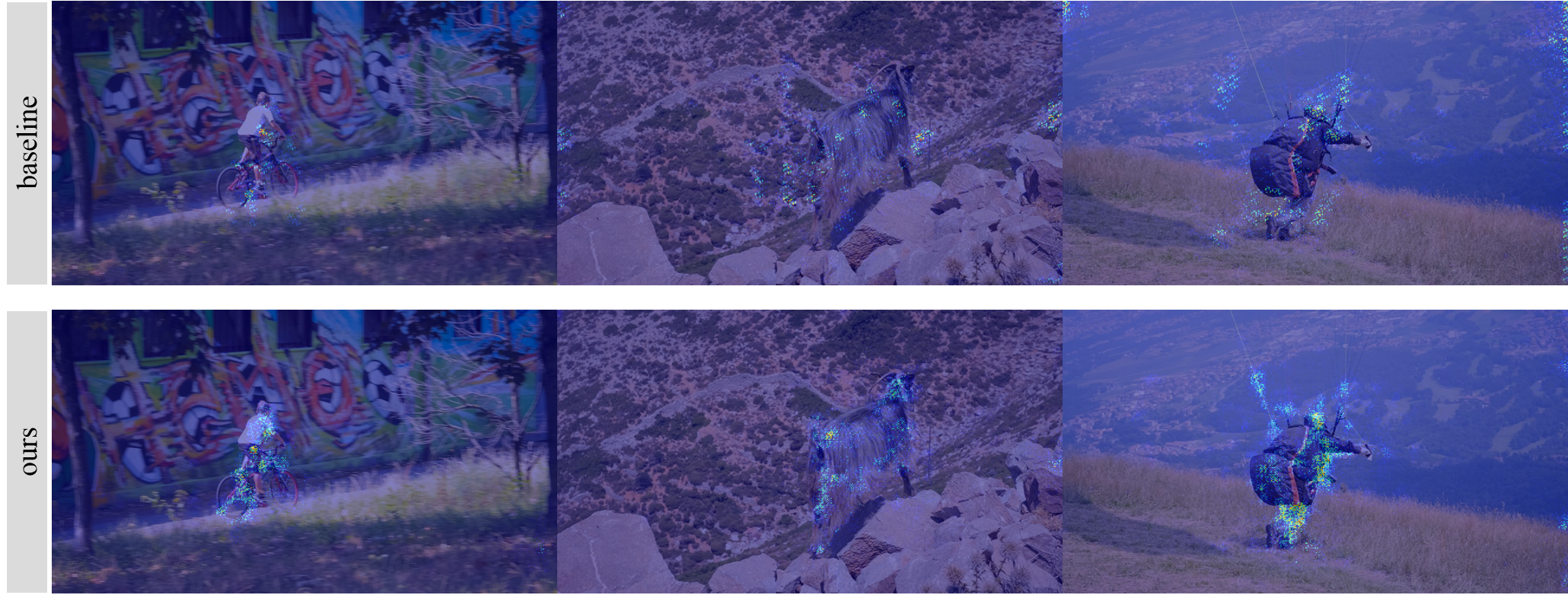}
    \caption{Cycle-ERF of frames w.r.t. the initial reference object masks in DAVIS17 validation.}
    \label{fig:erf}
\end{figure*}
\begin{figure*}
    \centering
    \includegraphics[width=0.95\textwidth]{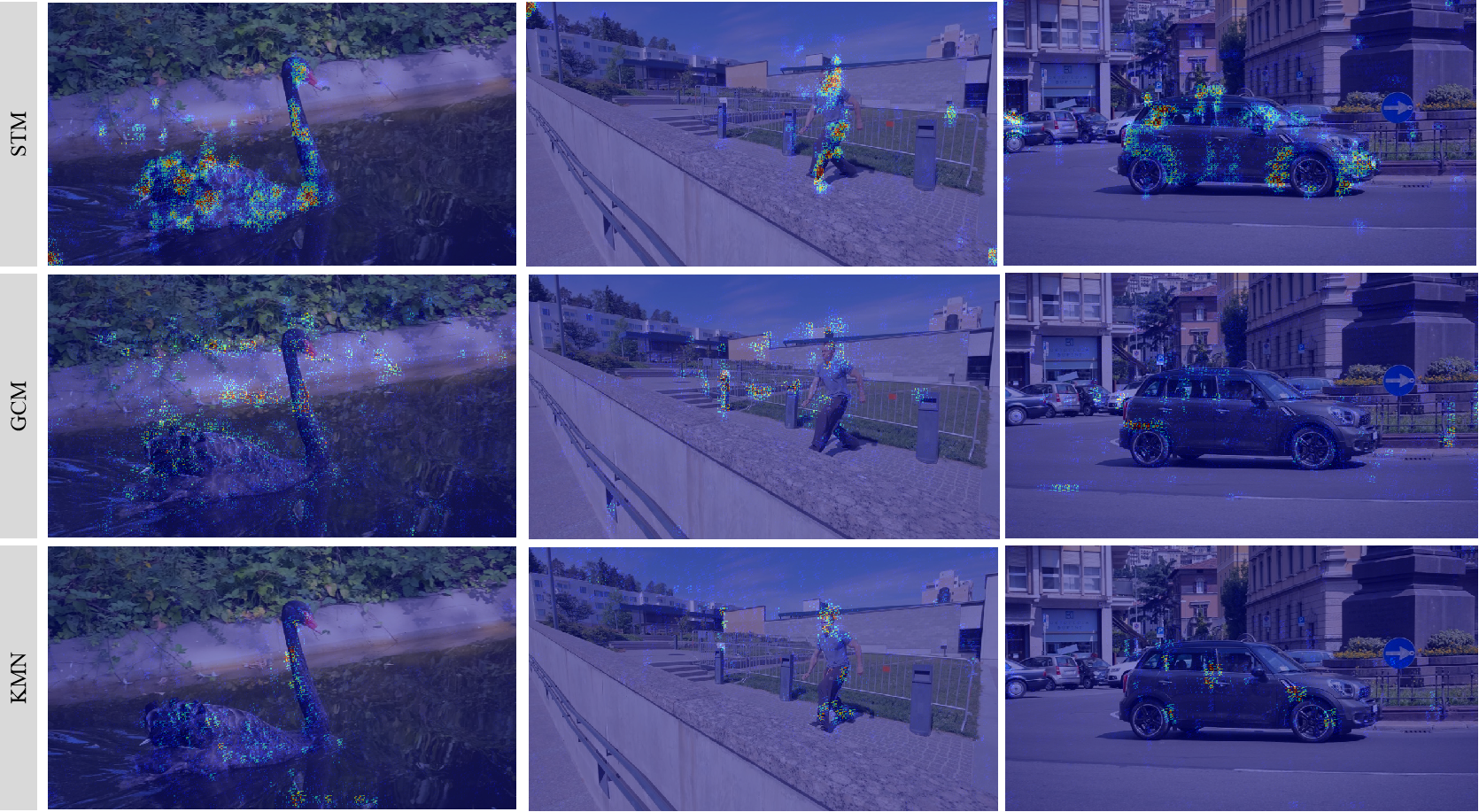}
    \caption{Cycle-ERF comparison on STM, GCM and KMN on DAVIS16 validation set.}
    \label{fig:model-erf}
\end{figure*}

\subsubsection{Extension to other memory-based methods.}

Our cyclic mechanism is easy to implement and can be naturally extend to other segmentation pipelines. To demonstrate the generalization of our design, we further extend another two famous baseline methods, KMN~\cite{KMN} GCM~\cite{GCVOS2020} \red{and AOT~\cite{AOT2021}} to a cyclic version and measure the performance gain. Corresponding results are reported in Table~\ref{tab:extension}, we can observe that combining with cyclic mechanism can boost the segmentation quality for all the baseline models on both datasets. In detail, we find the improvement of gradient correction on GCM is the least obvious, this can be due to the global property of GCM, where the memory squeezes the spatial domain, and loses the fine-grain information of original objects, which prevents the gradient from propagating to detailed spatial location. In contrast, the gain of gradient correction on KMN is most salient, since KMN itself will highlight the focusing area in the query frame, therefore the backward gradient related to the objects will be preserved. This is also reflected in the cycle-ERF analysis in section~\ref{sec:qualitative}.

\subsection{Qualitative analysis}\label{sec:qualitative}

\begin{figure*}
    \centering
    \includegraphics[width=0.96\textwidth]{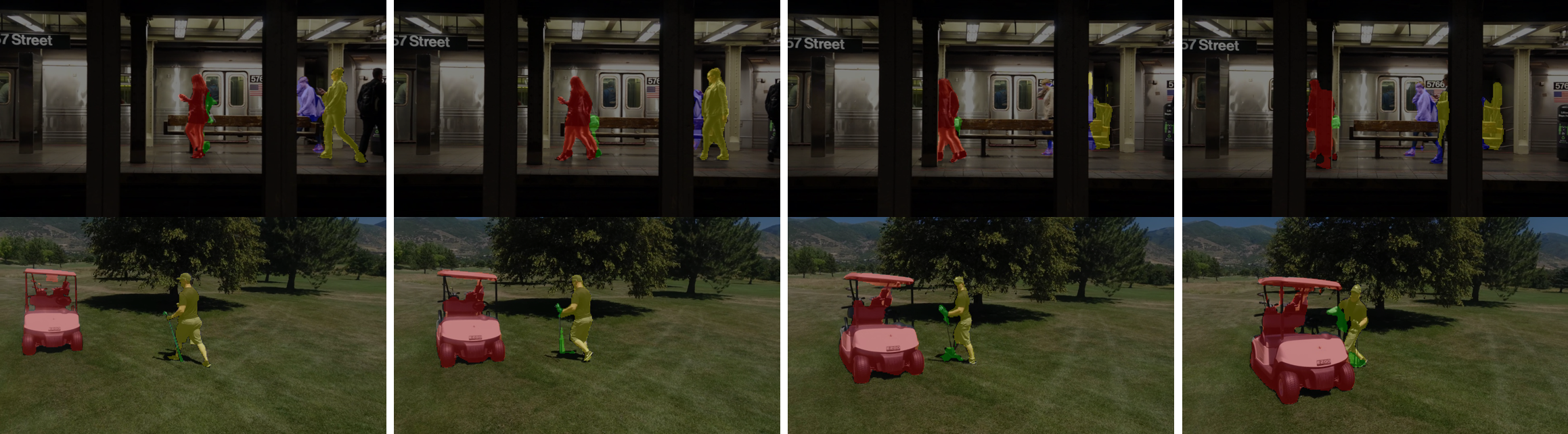}
    \caption{\red{Failure cases of STM-cycle on DAVIS17 test-dev.}}
    \label{fig:fail}
\end{figure*}

\noindent\textbf{Segmentation results.} In Figure~\ref{fig:vis}, we show some segmentation results using the STM model trained with and without our cycle scheme. From comparison on the first sequences, we observe that the cyclic mechanism suppresses the accumulative error from problematic reference masks. From the second video, we see the cyclic model can depicts the boundary between foreground objects more precisely, which is consistent with the quantitative results. Further, our method can successfully segment some challenging small objects (caught by the left woman's hand). In Figure~\ref{fig:correction}, we show the visual effect from the gradient correction module, it is clear to see that gradient correction can help update and reconstruct the losses on some detail.

\noindent\textbf{Cycle-ERF analysis.} We further analyze the cycle-ERF defined as Equation~(\ref{eq:erf}) on different approaches. We take the initial mask as the objects to be predicted and take a random intermediate frame and an empty mask as reference. Figure~\ref{fig:erf} visualizes the cycle-ERFs of some samples output from baseline STM and STM-cycle model. Compared with baseline, our cyclic training scheme helps the network concentrate more on the foreground objects with stronger responses. This indicates that our model learns more robust object-specific correspondence. It is also interesting to see that for STM network, only a small part of the objects is crucial for reconstructing the same objects that were in the initial frames as the receptive field focuses on the outline or skeleton of the objects. This can be used to explain the greater improvement of contour accuracy using our method, and also provide cues on the extraction from reference masks. 

In Figure~\ref{fig:model-erf}, we take the cycle-ERF as a tool to analysis the focusing area of different baseline models, STM~\cite{Oh_2019_ICCV}, GCM~\cite{GCVOS2020} and KMN~\cite{KMN} on DAVIS16. By comparison, we find STM shows overally stronger response on focusing area around the foreground object. In contrast, GCM highlight more background context around the object, since the memorial mechanism in this method always squeeze the context into the cache and focus more on global relationship. On the other hand, cycle-ERF from KMN yields response focusing on more specific and small part of objects and less intensity in background or context, this is in consistency with the design insight behind this method, where a dynamic gaussian kernel is applied to suppress the interaction between less related areas. These comparisons with Cycle-ERF manifest that such visualization methods can be a helpful tool to provide interpretability of existing models.

\red{\noindent\textbf{Investigation of Failure Cases.}  Figure~\ref{fig:fail} shows some failure cases of our method, although combined with cyclic loss and gradient correction, the network can not handle extremely narrow and small objects (e.g. the brassie in the man's hand in the second row), meanwhile, as shown in the first row, our method can suffer from cases where the specified objects are severely occluded by obstacles in the foreground.}

\section{Conclusion}
This paper incorporates the cycle mechanism with semi-supervised video segmentation network to mitigate the error propagation problem in current approaches. When combined with an efficient and flexible baseline, the proposed cyclic loss and gradient correction module achieve competitive performance-runtime trade-off on three challenging benchmarks. Detailed analysis are further conducted to demonstrate the generality and robustness of such cyclic design. Further explanations can be drawn from a new perspective of cycle-ERF.


%
%

\bibliographystyle{spmpsci}      
\bibliography{egbib}   

\begin{thebibliography}{10}
\providecommand{\url}[1]{{#1}}
\providecommand{\urlprefix}{URL }
\expandafter\ifx\csname urlstyle\endcsname\relax
  \providecommand{\doi}[1]{DOI~\discretionary{}{}{}#1}\else
  \providecommand{\doi}{DOI~\discretionary{}{}{}\begingroup
  \urlstyle{rm}\Url}\fi

\bibitem{stm-coco}
Pretraining code of space-time-memory network on coco for video object
  segmentation.
\newblock \url{https://github.com/haochenheheda/Training-Code-of-STM}

\bibitem{Bansal2018Recycle}
Bansal, A., Ma, S., Ramanan, D., Sheikh, Y.: Recycle-gan: Unsupervised video
  retargeting.
\newblock In: European Conference on Computer Vision (ECCV) (2018)

\bibitem{Cae_OVOS_17}
Caelles, S., Maninis, K.K., Pont-Tuset, J., Leal-Taix\'e, L., Cremers, D., {Van
  Gool}, L.: One-shot video object segmentation.
\newblock In: Computer Vision and Pattern Recognition (CVPR) (2017)

\bibitem{trackcolor}
Carl, V., Abhinav, S., Alireza, F., Sergio, G., Kevin, M.: Tracking emerges by
  colorizing videos.
\newblock European Conference on Computer Vision  (2018)

\bibitem{dong2018boosting}
Dong, Y., Liao, F., Pang, T., Su, H., Zhu, J., Hu, X., Li, J.: Boosting
  adversarial attacks with momentum.
\newblock In: Proceedings of the IEEE conference on computer vision and pattern
  recognition (CVPR), pp. 9185--9193 (2018)

\bibitem{Everingham15}
Everingham, M., Eslami, S.M.A., Van~Gool, L., Williams, C.K.I., Winn, J.,
  Zisserman, A.: The pascal visual object classes challenge: A retrospective.
\newblock International Journal of Computer Vision (IJCV) \textbf{111}(1),
  98--136 (2015)

\bibitem{goodfellow2014explaining}
Goodfellow, I.J., Shlens, J., Szegedy, C.: Explaining and harnessing
  adversarial examples.
\newblock arXiv preprint arXiv:1412.6572  (2014)

\bibitem{He2018Mask}
He, K., Georgia, G., Piotr, D., Ross, G.: Mask r-cnn.
\newblock In: International Conference on Computer Vision (ICCV) (2018)

\bibitem{He_2016_CVPR}
He, K., Zhang, X., Ren, S., Sun, J.: Deep residual learning for image
  recognition.
\newblock In: The IEEE Conference on Computer Vision and Pattern Recognition
  (CVPR) (2016)

\bibitem{jabri2020walk}
Jabri, A., Owens, A., Efros, A.A.: Space-time correspondence as a contrastive
  random walk.
\newblock Advances in Neural Information Processing Systems  (2020)

\bibitem{agame}
Johnander, J., Danelljan, M., Brissman, E., Khan, F.S., Felsberg, M.: A
  generative appearance model for end-to-end video object segmentation.
\newblock In: IEEE/CVF Conference on Computer Vision and Pattern Recognition
  (CVPR), pp. 8953--8962 (2019)

\bibitem{lucid_dreaming}
Khoreva, A., Benenson, R., Ilg, E., Brox, T., Schiele, B.: Lucid data dreaming
  for video object segmentation.
\newblock International Journal of Computer Vision (IJCV) \textbf{127},
  1175–1197 (2019)

\bibitem{adam2014method}
Kingma, D.P., Ba, J.: Adam: A method for stochastic optimization.
\newblock In: International Conference on Learning Representation (ICLR) (2014)

\bibitem{GCVOS2020}
Li, Y., Shen, Z., Shan, Y.: Fast video object segmentation using the global
  context module.
\newblock In: European Conference on Computer Vision (ECCV), pp. 735--750
  (2020)

\bibitem{Li_2020_NeurIPS}
Li, Y., Xu, N., Jinlong, P., See, J., Weiyao, L.: Delving into the cyclic
  mechanism in semi-supervised video object segmentation.
\newblock In: Neural Information Processing System (NeurIPS) (2020)

\bibitem{Lin_2019_ICCV}
Lin, H., Qi, X., Jia, J.: Agss-vos: Attention guided single-shot video object
  segmentation.
\newblock In: The IEEE International Conference on Computer Vision (ICCV)
  (2019)

\bibitem{COCO}
Lin, T.Y., Maire, M., Belongie, S., Hays, J., Perona, P., Ramanan, D., Dollár,
  P., Zitnick, C.L.: Microsoft coco: Common objects in context.
\newblock In: European Conference on Computer Vision (ECCV). Zürich (2014)

\bibitem{luiten2018premvos}
Luiten, J., Voigtlaender, P., Leibe, B.: Premvos: Proposal-generation,
  refinement and merging for video object segmentation.
\newblock In: Asian Conference on Computer Vision (ACCV) (2018)

\bibitem{Meister:2018:UUL}
Meister, S., Hur, J., Roth, S.: {UnFlow}: Unsupervised learning of optical flow
  with a bidirectional census loss.
\newblock In: AAAI. New Orleans, Louisiana (2018)

\bibitem{Oh_2019_ICCV}
Oh, S.W., Lee, J.Y., Xu, N., Kim, S.J.: Video object segmentation using
  space-time memory networks.
\newblock In: The IEEE International Conference on Computer Vision (ICCV)
  (2019)

\bibitem{peng2020ctracker}
Peng, J., Wang, C., Wan, F., Wu, Y., Wang, Y., Tai, Y., Wang, C., Li, J.,
  Huang, F., Fu, Y.: Chained-tracker: Chaining paired attentive regression
  results for end-to-end joint multiple-object detection and tracking.
\newblock In: The European Conference on Computer Vision (ECCV) (2020)

\bibitem{Perazzi_2017_CVPR}
Perazzi, F., Khoreva, A., Benenson, R., Schiele, B., Sorkine-Hornung, A.:
  Learning video object segmentation from static images.
\newblock In: The IEEE Conference on Computer Vision and Pattern Recognition
  (CVPR) (2017)

\bibitem{Perazzi2016}
Perazzi, F., Pont-Tuset, J., McWilliams, B., {Van Gool}, L., Gross, M.,
  Sorkine-Hornung, A.: A benchmark dataset and evaluation methodology for video
  object segmentation.
\newblock In: Computer Vision and Pattern Recognition (CVPR) (2016)

\bibitem{Pont-Tuset_arXiv_2017}
Pont-Tuset, J., Perazzi, F., Caelles, S., Arbel\'aez, P., Sorkine-Hornung, A.,
  {Van Gool}, L.: The 2017 davis challenge on video object segmentation.
\newblock arXiv:1704.00675  (2017)

\bibitem{Robinson_2020_CVPR}
Robinson, A., Lawin, F.J., Danelljan, M., Khan, F.S., Felsberg, M.: Learning
  fast and robust target models for video object segmentation.
\newblock In: IEEE/CVF Conference on Computer Vision and Pattern Recognition
  (CVPR) (2020)

\bibitem{imagenet_cvpr09}
Russakovsky, O., Deng, J., Su, H., Krause, J., Satheesh, S., Ma, S., Huang, Z.,
  Karpathy, A., Khosla, A., Bernstein, M., Berg, A.C., Fei-Fei, L.: Imagenet
  large scale visual recognition challenge.
\newblock International Journal of Computer Vision (IJCV) \textbf{115}(3),
  211–252 (2015)

\bibitem{KMN}
Seong, H., Hyun, J., Kim, E.: Kernelized memory network for video object
  segmentation.
\newblock In: European Conference on Computer Vision (ECCV), pp. 629--645
  (2020)

\bibitem{CSSD}
{Shi}, J., {Yan}, Q., {Xu}, L., {Jia}, J.: Hierarchical image saliency
  detection on extended cssd.
\newblock IEEE Transactions on Pattern Analysis and Machine Intelligence
  (TPAMI) \textbf{38}(4), 717--729 (2016)

\bibitem{cycle3d}
Tinghui, Z., Philipp, K., Mathieu, A., Qixing, H., Alexei, A.E.: Learning dense
  correspondence via 3d-guided cycle consistency.
\newblock The IEEE Conference on Computer Vision and Pattern Recognition (CVPR)
   (2016)

\bibitem{Ventura_2019_CVPR}
Ventura, C., Bellver, M., Girbau, A., Salvador, A., Marques, F., Giro-i Nieto,
  X.: Rvos: End-to-end recurrent network for video object segmentation.
\newblock In: The IEEE Conference on Computer Vision and Pattern Recognition
  (CVPR) (2019)

\bibitem{Voigtlaender_2019_CVPR}
Voigtlaender, P., Chai, Y., Schroff, F., Adam, H., Leibe, B., Chen, L.C.:
  Feelvos: Fast end-to-end embedding learning for video object segmentation.
\newblock In: The IEEE Conference on Computer Vision and Pattern Recognition
  (CVPR) (2019)

\bibitem{voigtlaender17BMVC}
Voigtlaender, P., Leibe, B.: Online adaptation of convolutional neural networks
  for video object segmentation.
\newblock In: British Machine Vision Conference (BMVC) (2017)

\bibitem{CVPR2019_CycleTime}
Wang, X., Jabri, A., Efros, A.A.: Learning correspondence from the
  cycle-consistency of time.
\newblock In: The IEEE Conference on Computer Vision and Pattern Recognition
  (CVPR) (2019)

\bibitem{Oh_2018_CVPR}
Wug~Oh, S., Lee, J.Y., Sunkavalli, K., Joo~Kim, S.: Fast video object
  segmentation by reference-guided mask propagation.
\newblock In: The IEEE Conference on Computer Vision and Pattern Recognition
  (CVPR) (2018)

\bibitem{Xu_2018_S2S_ECCV}
Xu, N., Yang, L., Fan, Y., Yang, J., Yue, D., Liang, Y., Price, B., Cohen, S.,
  Huang, T.: Youtube-vos: Sequence-to-sequence video object segmentation.
\newblock In: The European Conference on Computer Vision (ECCV) (2018)

\bibitem{AOT2021}
Yang, Z., Wei, Y., Yang, Y.: Associating objects with transformers for video
  object segmentation.
\newblock In: Advances in Neural Information Processing Systems, vol.~34, pp.
  2491--2502 (2021)

\bibitem{Zeng_2019_ICCV}
Zeng, X., Liao, R., Gu, L., Xiong, Y., Fidler, S., Urtasun, R.: Dmm-net:
  Differentiable mask-matching network for video object segmentation.
\newblock In: The IEEE International Conference on Computer Vision (ICCV)
  (2019)

\bibitem{zhang2020a}
Zhang, Y., Wu, Z., Peng, H., Lin, S.: A transductive approach for video object
  segmentation.
\newblock In: Proceedings of the IEEE Conference on Computer Vision and Pattern
  Recognition (2020)

\bibitem{CycleGAN2017}
Zhu, J.Y., Park, T., Isola, P., Efros, A.A.: Unpaired image-to-image
  translation using cycle-consistent adversarial networks.
\newblock In: IEEE International Conference on Computer Vision (ICCV) (2017)

\end{thebibliography}


\end{document}